# A fuzzy adaptive metaheuristic algorithm for identifying sustainable, economical, lightweight, and earthquake-resistant reinforced concrete cantilever retaining walls

Farshid Keivanian[1], Raymond Chiong[1,*], Ali R. Kashani[2], Amir H. Gandomi[3]

[1]School of Information and Physical Sciences, The University of Newcastle, Callaghan, NSW 2308, Australia

[2]Department of Civil Engineering, University of Memphis, Memphis, TN 38152, USA

[3]Faculty of Engineering and Information Technology, University of Technology Sydney, Ultimo, NSW 2007, Australia

**ABSTRACT**

In earthquake-prone zones, the seismic performance of reinforced concrete cantilever (RCC) retaining walls is significant. In this study, the seismic performance was investigated using horizontal and vertical pseudo-static coefficients. To tackle RCC weights and forces resulting from these earth pressures, 26 constraints for structural strengths and geotechnical stability along with 12 geometric variables are associated with each design. These constraints and design variables form a constraint optimization problem with a twelve-dimensional solution space. To conduct effective search and produce sustainable, economical, lightweight RCC designs robust against earthquake hazards, a novel adaptive fuzzy-based metaheuristic algorithm is applied. The proposed method divides the search space to sub-regions and establishes exploration, information sharing, and exploitation search capabilities based on its novel search components. Further, fuzzy inference systems were employed to address parameterization and computational cost evaluation issues. It was found that the proposed algorithm can achieve low-cost, low-weight, and low-$CO_2$ emission RCC designs under nine seismic conditions in comparison with several classical and best-performing design optimizers.

## 1 Introduction

Retaining wall structures address the instability issues of soil trenches (natural and artificial). These structures are divided into four types: gravity, counterfort, cantilever, and anchored retaining walls [1]. The reinforced concrete cantilever (RCC) retaining walls, in particular, are designed to resist against a combination of earth and hydrostatic loading (i.e. lateral soil pressure) in road construction on railways, highways, bridges, etc., where there is a change in ground elevation [2]. Considering their wide application and massive construction volume, the optimal design of RCC walls is essential. However, a significant number of independent design variables increase the size of search domain exponentially [3], while the highly nonlinear constraints involved in the design procedure intensify the complexity of this task. Hence, automated optimization methods may offer an efficient solution to this issue.

---
\* Corresponding author's email: Raymond.Chiong@newcastle.edu.au

Optimization algorithms can be broadly divided into two main categories: mathematical programming (i.e. exact methods) and metaheuristic optimization methods. Classical methods attempt to calculate optimal solutions via iterative linear or constrained nonlinear programming techniques in a deterministic way [4-9]. As an alternative, metaheuristics with inherent stochastic and evolutionary search features have gained wide attention to leave more room for uncertainty levels in modelling optimization problems with design variables [10-12]. The primary advantage of metaheuristics is their efficient use of the "trial-and-error" principle in searching for solutions. To be more exact, metaheuristics try to get as close to the optimal solution as possible rather than finding the exact extrema. Owing to this strategy, metaheuristics have been successfully applied to solve various complex optimization problems.

Metaheuristics can be grouped into three broad clusters: 1) Swarm Intelligence (SI), 2) Evolutionary Algorithms (EAs), and 3) Physical System (PS)-inspired methods [13]. SI mimics the collective intelligence behavior of self-organized simple agents without any centralized control. SI-based algorithms include the Particle Swarm Optimization (PSO) [14], Accelerated PSO (APSO) [15], Ant Colony Optimization (ACO) [16], Imperialist Competitive Algorithm (ICA) [17], Cuckoo Search (CS) [18], Firefly Algorithm (FA) [19], Flower Pollination Algorithm (FPA) [19], Adaptive Dimensional Search technique (ADS) [20], and Grey Wolf Optimization (GWO) [21]. EAs imitate the biological evolutionary processes, such as reproduction, mutation, recombination, and selection in nature, to perform optimization. Evolution of the population takes place after repeated applications of the above operators. EA-based representatives are the Evolutionary Strategy (ES) [22], Genetic Algorithm (GA) [23], Differential Evolution (DE) [24], Search Group Algorithm (SGA) [25], Backtracking Search Algorithm (BSA) [26], Big Bang-Big Crunch optimization (BB-BC) [27], Biogeography-Based Optimization (BBO) [28], BBO with Levy flight distribution (LFBBO) [29], and Interior Search Algorithm (ISA) [30]. While SI like PSO is based on the behaviors of birds and fish [31], other metaheuristics have been derived from the rules of physics in the universe. The Harmony Search (HS) algorithm [32], inspired by the underlying principles of musicians' improvisation of harmony, consists of three operators: random search, harmony memory considering rule, and pitch adjusting rule to handle exploration and exploitation. Another PS-inspired example is Simulated Annealing (SA) [33], which is based on the annealing process of metals.

Due to the complex nature of civil engineering problems, in particular, geotechnical engineering-related problems, such as RCC retaining wall design optimization problems, suffer from a high level of dimensionality, uncertainties caused by a large number of decision variables, high non-linearity caused by constraints. Therefore, the number of studies to develop a robust metaheuristic optimizer with more efficient search capabilities has followed an increasing trend. Recently, researchers have comprehensively investigated the efficiency of the standard SI, EA, and PS-inspired exemplars and their modified variants on solving different RCC wall design issues [20, 34]. For instance, improved HS (IHS) [35, 36] and modified SA (MSA) [37] have been proposed using adaptive control parameters, and LFBBO [29, 38] was introduced using a Levy flight distribution index incorporated into BBO's mutation policy for better performance [39, 40]. Kashani et al. [39] reviewed the most recent EA-based efforts, such as BBO and LFBBO, and SI-based exemplars like the ISA developed to handle the optimum design of RCCs with a large number of decision variables between 2011 and 2018. Kashani et al. [40] also evaluated six variants of the most common SI-based approach, the PSO and its variations devoted to handling various RCC designs between 2006-2015. This review study clearly demonstrated that most of the PSO variants are capable of solving RCC wall designs.

Other SI-based methods, such as GWO and the FPA, have been adopted, for the first time, to determine the optimum weight and cost design of RCC retaining walls quite recently. Comparing several conventional SI methods (i.e. ISA, PSO, APSO, FA, and CS) and EAs (i.e. SGA, BSA, BBO, and BB-BC), GWO was found to be the most efficient method in saving total weights [41]. In GWO, the sum of the weights of concrete and steel reinforcement materials are formulated as its objective function. Considering ACI 318-05 (2005) codes [42], 12 discrete-continuous design variables (8 continuous geometric design variables related to dimensions of the wall and 4 discrete structural design variables associated with reinforcement) were set. Moreover, 26 constraints (3 geotechnical constraints as the factor of safety against overturning, bending, bearing capacity failure and 23 structural constraints as the moment and shear capacities of for elements of the wall) were associated to each design. The FPA outperformed a standard GA and PSO, yielding lesser cost and smaller variability [2], in which the sum of the costs of concrete and reinforcing steel materials and soil excavation was set as the objective function. Six continuous design variables were considered to describe the geometry of the wall, while structural design parameters were assumed fixed. The constraints were set in accordance with Eurocode 7 (EC7) [43].

Dynamic analysis is a step toward addressing structural design efficiency against earthquake hazards, which has attracted the focus of many studies on RCC structures [44-46]. The analysis of an RCC under seismic loading is a complicated soil-structure interaction task [38]; therefore, the so-called 'pseudo-static' technique is applied to simulate and simplify the real behavior. This technique is based on static analysis, applying equivalent horizontal and vertical seismic coefficients. The seismic design consideration has been addressed by the limited number of geotechnical expert communities [38, 47-49]. In a recent study, Gandomi and Kashani [49] examined the optimum design of RCCs, each affected by nine combinations of both horizontal and vertical pseudo-static loading cases. The constraints applied to the design procedure intensify the non-linear complexity level of the solution space. Sustainable development of construction projects is in-demand due to increasing climate change in the 21$^{st}$ century. The environmental impacts of RCCs involve embodied emissions of reinforcing steel and concrete construction [50], whereby reinforcing steel causes embodied emission due to the energy used to melt and reform scrap metal [51]. The concrete contributes embodied emissions caused by fuel combustion and carbon oxidation during clinker and cement production, which is responsible for roughly 8% of global $CO_2$ emissions [52]. Embodied $CO_2$ emissions can be reduced by using structural optimization methodologies for maximizing material efficiency and minimizing the environmental impact of RCC structures. A related study set the life-cycle cost and environmental impact as design objectives of RCC design and indicated the close relation between embodied emissions and total cost.

The presented study aims to fill the research gap in reducing the environmental impacts of concrete and reinforcing steel on the properties of the optimum seismic designs of RCCs by conducting a novel and robust automated optimization technique that maximizes structural efficiency. Herein, for the first time, a novel adaptive fuzzy-based metaheuristic approach was applied for automating pseudo-static analysis of sustainable, economical, lightweight RCC walls by determining 12 design variables (8 geometry and 4 steel reinforcement variables) and setting 3 types of objective functions: low-cost, low-weight, and low-$CO_2$ design. This novel fuzzy-based approach was inspired by the findings from our recent fuzzy logic-based metaheuristic framework on solving optimization benchmark functions of different dimensions [53] and suggestions from the literature to monitor every succeeding run at certain intervals [34] for a satisfactory trade-off between exploration and exploitation features of the search process [20]. The main goals here are: 1) to identify efficient design practices that minimize the environmental impact of earthquake-resistant RCC designs, and 2) to examine the trade-offs between the objectives that will be achieved by conducting novel search strategies.

The proposed approach splits the search space into sub-populations and incorporates our novel search strategies to each sub-population, which helps to boost exploration, exploitation, and information sharing capabilities of the approach, avoiding premature convergence when locating the optimal set of design variables. Furthermore, we employed fuzzy inference systems (FISs) to address parameterization and computational cost evaluation issues, which help to balance the search capabilities more efficiently. Fuzzy inferencing acts as both a learning strategy adaptor in each run and a smart operator selector in each running window. A computer program based on ACI 318-05 (2005) [42] requirements for structural strengths and DAS (1994) [54] criteria for the geotechnical stability was developed using MATLAB. To evaluate the efficiency of the final optimum designs against horizontal earth pressures, two numerical examples, each affected by the nine combinations of horizontal and vertical seismic coefficients between 0 and 0.3, were conducted. Statistical analysis based on the Wilcoxon rank-sum test is used to verify the statistical significance of our experimental results.

The rest of this paper is organized as follows. Section 2 describes the problem formulation. Section 3 defines the objective functions. Section 4 discusses the research gaps. Section 5 introduces the fuzzy set theory in detail. Section 6 presents the proposed algorithm. Section 7 provides the simulation results, numerical analysis and discussions. Section 8 concludes the work and outlines future research plans. Finally, Appendix A and Appendix B show convergence histories, and Appendix C includes detailed statistical analysis.

## 2 Problem formulation

A schematic view of a retaining wall is depicted in Fig. 1. The wall is defined by eight geometrical design variables $X_i$ ($i$=1 to 8) and four reinforcement design variables $R_i$ ($i$=1 to 4): $X_1$ is the width of base slab; $X_2$ is the width of toe slab; $X_3$ and $X_4$ are the stem thickness at the bottom of and top of the wall, respectively; $X_5$ is the base slab thickness; $X_6$ is the distance from the front of the sear key to the front of the tow of the wall; $X_7$ is the width of the shear key; $X_8$ is the height of shear key; $R_1$ is the steel area of the stem, $R_2$ is the steel area of the toe slab; $R_3$ is the steel area of the heel slab; and $R_4$ is the steel area of the shear key. Steel reinforcement areas are considered to be discrete variables selected from a design pool proposed by Camp and Akin [55].

Fig. 2 shows the effective loads acting on retaining wall, including the total weight of the retaining wall $W$ (i.e. the weight concrete wall $W_c$, the weight of backfill acting on the heel of the wall $W_s$, the weight of soil on the toe of the wall $W_T$), the distributed surcharge load $q$, the force resulting from the active earth pressure $P_a$, the force resulting from passive earth pressure on the base shear key $P_k$, the force resulting from passive earth pressure on the front part of the toe section of the wall $P_p$, and the force resulting from the bearing stress of the base soil $P_B$. The resultant force, $R$, of the normal and resisting shear forces will be inclined at angle $\varphi$.

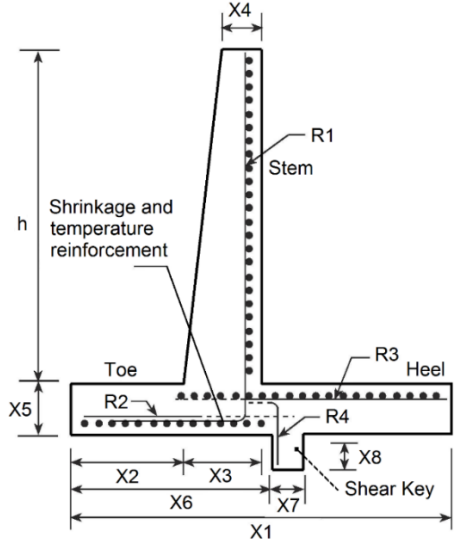

Fig. 1: Design variables for retaining wall design problem [49]

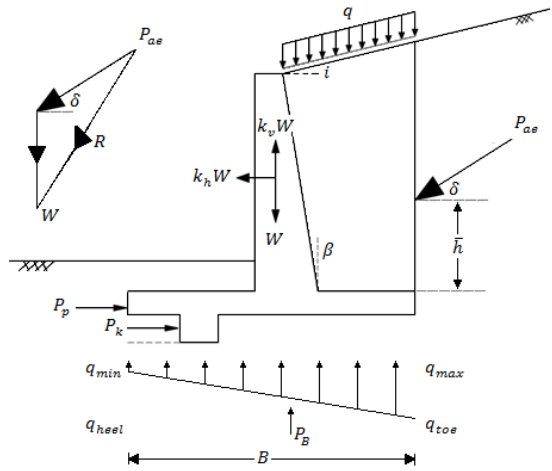

Fig. 2: General forces acting on a reinforced cantilever retaining wall

The analysis of a retaining wall under seismic loads is especially critical in seismic zones. However, determining the real behavior of retaining walls under seismic loads is a complex soil-structure interaction problem and can be impeded by differing opinions, diverse theories, and site tests that are not consistent with the theories [38]. For these reasons, simplified 'pseudo-static' approaches are developed to predict the actual dynamic behavior of the retaining wall. In the pseudo-static approaches, the seismic loads are converted to equivalent static forces, which enables the analyses of the retaining walls in the static case.

One of the most common pseudo-static approaches is the Mononobe-Okabe (M-O) theory, proposed by Mononobe and Okabe in the 1920s [56], [57]. The theory, developed for dry, cohesion-less soils, is based on three assumptions [58]: 1) the retaining wall can move sufficiently to develop minimum active pressure; 2) by the time the minimum $P_a$ value is reached, the region behind a point of soil wedge where the initial failure and the maximum shear stress occurred moves along the potential sliding surface; and 3) the back soil behaves as a rigid body. Therefore, accelerations have uniform distribution from the beginning to the end of the wall. Also, Coulomb's theory for active earth pressure refers to the forces caused by an earthquake, by which the wall moves away from the soil mass to the left in Fig. 2 [59]. Therefore, there are two forces, $k_h W$ and $k_v W$, in the horizontal and vertical directions ($k_h$ and $k_v$ are the horizontal and vertical coefficients based on geographic characteristics of the site, respectively).

In the M-O theory, total maximum active earth force per unit length of the wall, $P_{ae}$, is evaluated by:

$$P_{ae} = \frac{1}{2} k_{ae} \gamma_{soil} h^2 (1 - k_v) \tag{1}$$

where $\gamma_{soil}$ is the soil density, which measures the unit weight of soi; $h$ is the height of wall; $k_v$ is the vertical acceleration coefficient; and $k_{ae}$ is the dynamic active earth pressure coefficient that is computed by:

$$k_{ae} = \frac{\cos^2(\emptyset - \theta - \beta)}{\cos\theta \cos^2\beta \cos(\delta + \theta + \beta)\left[1 + \sqrt{\frac{\sin(\delta + \emptyset)\sin(\emptyset - \theta - i)}{\cos(\delta + \beta + \theta)\cos(i - \beta)}}\right]^2} \tag{2}$$

where $\delta$ refers to the angle of wall friction between soil and rear face of the stem; $\emptyset$ is the angle of internal friction of the soil; the angle $\beta$ is the slope of the back wall with respect to the vertical, backfill, and ground behind the wall; $i$ is the slope of the ground surface behind the wall; and $\theta$ is the seismic inertia angle, which measures the friction angle of the backfill slope by:

$$\theta = \tan^{-1}\left(\frac{k_h}{1 - k_v}\right) \tag{3}$$

where $k_h$ is the horizontal earthquake acceleration coefficient. In this study, horizontal and vertical coefficients take nine distinct combinations from 0 to 0.3 in increments of 0.15.

It is mentioned in the M-O theory that $P_{ae}$ acts at the point that is $h/3$ above the base. However, according to experimental studies [38], this point is higher than the actual acting points of the total active force under seismic loads. Therefore, $P_{ae}$ can be divided into two components: the static $P_a$ and dynamic $\Delta P_{ae}$, as follows:

$$P_{ae} = P_a + \Delta P_{ae} \tag{4}$$

The static component, according to Coulomb [60], is described by:

$$P_a = \frac{1}{2} k_a \gamma_{soil} h^2 \tag{5}$$

$$k_a = \frac{\cos^2(\emptyset - \theta)}{\cos^2\theta \cos(\delta + \theta)\left[1 + \sqrt{\frac{\sin(\delta + \emptyset)\sin(\emptyset - \beta)}{\cos(\delta + \theta)\cos(\beta - \theta)}}\right]^2} \tag{6}$$

According to Seed and Whitman [61], the acting point of the dynamic component is almost $0.6h$ above the base, where $\bar{h}$ is the height between the base and the acting point of the total active force:

$$\bar{h} = \frac{P_a(h/3) + \Delta P_{ae}(0.6h)}{P_{ae}} \tag{7}$$

Passive earth pressure occurs when the wall moves into the soil mass and is expressed similar to that discussed for active earth pressure per unit length of the wall:

$$P_{pe} = \frac{1}{2} k_{pe} \gamma_{soil} h^2 (1 - k_v) \tag{8}$$

where $k_{pe}$ is the dynamic passive earth pressure coefficient and is expressed as:

$$k_{pe} = \frac{\cos^2(\emptyset + \theta - \beta)}{\cos\theta \cos^2\beta \cos(\delta - \theta + \beta) \left[1 + \sqrt{\frac{\sin(\emptyset - \delta)\sin(\emptyset + i - \theta)}{\cos(\delta - \beta + \theta)\cos(i - \beta)}}\right]^2} \tag{9}$$

Similarly, the total passive force can be divided into static $P_p$ and dynamic $\Delta P_{pe}$ components. Using Coulomb's theory [60], the static component is described as follows:

$$P_p = \frac{1}{2} k_p \gamma_{soil} h^2 \tag{10}$$

$$k_p = \frac{\cos^2(\emptyset + \theta)}{\cos^2\theta \cos(\delta - \theta) \left[1 + \sqrt{\frac{\sin(\emptyset + \delta)\sin(\emptyset + \beta)}{\cos(\delta - \theta)\cos(\beta + \theta)}}\right]^2} \tag{11}$$

Furthermore, $\Delta P_{pe}$ acts in the opposite direction of $P_p$, meaning that the remaining passive force is reduced.

The problem was resolved for the nine different active and passive earth pressure conditions, each representing a combination of pseudo-static coefficients. The schematic of a retaining wall affected by seismic loading cases along with all possible effective forces is presented in Fig. 2. The RCC cantilever has a vertical wall of height *h*, which retains soil, as shown in Fig. 2. A uniformly distributed surcharge load, *q*/unit area is also applied at the ground surface. As the wall may tilt away from the soil that is retained, the lateral pressure for this condition is referred to as *active earth pressure*. On the other side, the wall may be pushed into the soil that is retained, thus the lateral pressure for this condition is referred to as *passive earth pressure* on the base shear key [59].

The constraints of the optimization problem are determined according to the requirements for the geotechnical, structural, and geometrical design of the RCC walls. The design of these walls is based on three basic analyses: (1) geotechnical stability, (2) structural stability, and (3) geometrical analyses. The geotechnical stability analysis phase, as specified by DAS [59], consists of three checks: overturning, sliding, and bearing capacity stability checks. The overturning check is realized by calculating the overturning factor of safety, $FS_O$:

$$FS_O = \frac{\sum M_R}{\sum M_O} \tag{12}$$

where $\sum M_R$ is the sum of resisting moment against overturning; and $\sum M_O$ is the sum of overturning moments. To satisfy this criterion, $FS_O \geq 1.5$.

In a similar manner, the sliding check is conducted by evaluating the sliding factor of safety, $FS_S$:

$$FS_S = \frac{\sum F_R}{\sum F_D} = \frac{\left(\sum W_{wall} \tan\left(\frac{2\emptyset_{base}}{3}\right)\right) + \frac{2Bc_{base}}{3} + P_p + P_k}{P_a \cos\beta} \tag{13}$$

where $\sum F_R$ is the sum of horizontal resisting forces against sliding; $\sum F_D$ is the sum of horizontal driving forces; $\sum W_{wall}$ is the total weight of the retaining wall; $\emptyset_{base}$ is the internal friction of the base soil; $B$ is the total width of the base slab; $c_{base}$ is the adhesion between the soil and base slab; $P_p$ is the passive force in front of the base slab; $P_k$ is the force in front of the shear key; and $P_a$ is the active force. To satisfy this criterion, $FS_S \geq 1.5$.

In the next step of the wall design procedure, the bearing stress underneath the retaining walls is evaluated as the bearing factor of safety, $FS_B$:

$$FS_B = \frac{1.33 q_a}{q_{max}} \tag{14}$$

where $q_a$ is the soil foundation ultimate bearing capacity and $q_{max}$ is the maximum stress acting along the base of the structure. In Eq. (14), the value of $q_a$ employed for static loading designs increases by 33% for seismic loading conditions, according to AASHTO [62]. The maximum and minimum stresses acting along the base of the retaining wall can be determined as follows:

$$q_{\substack{min \\ max}} = \frac{\sum V}{B}\left(1 \mp \frac{6e}{B}\right) \tag{15}$$

$$e = \frac{B}{2} - \frac{\sum M_R - \sum M_O}{\sum V} \tag{16}$$

where $\sum V$ is sum of all the vertical forces (consequent 8 forces of the wall, weight of the soil above the base, and the surcharge load); and $e$ is the eccentricity of the consequence force system. To satisfy the bearing strength criterion, $FS_B \geq 3$.

In the second phase of the wall design, the structural stability, as specified in ACI 318-05 [42], is checked. To end this, all sections of the wall, including the stem, heel, tow, and shear key sections of the wall, are checked for shear and moment capacity based on (17) and (18), respectively:

$$\frac{M_n}{M_u} \geq 1 \tag{17}$$

$$\frac{V_n}{V_u} \geq 1 \tag{18}$$

where $V_n$ and $M_n$ are the shear and flexural strengths assessed by (19) and (20), respectively; and $V_u$ and $M_u$ are the factored shear and flexural demands, respectively.

$$V_n = \emptyset_v \, 0.17 \sqrt{f_c} bd \tag{19}$$

$$M_n = \emptyset_m A_s f_y \left(d - \frac{a}{2}\right) \tag{20}$$

where $\emptyset_v = 0.75$ and $\emptyset_m = 0.9$ (ACI 318-05 [42]); $A_s$ is the area of the steel reinforcement; $f_y$ is the yield stress of the steel reinforcement; $d$ is the distance from the compression surface to the centroid of the tension steel; $a$ is the depth of the stress block; and $b$ is the width of the section.

$V_u$ and $M_u$ are the final shear force and moment, respectively, for which the design moment and shear force for the tow and heel slabs will be computed using Eqs. (21)-(24). The effective design forces on the base slab are depicted in Fig. 2.

$$M_t = \left[1.7\left(\frac{q_2}{6} + \frac{q_{max}}{3}\right) - 0.9(\gamma_c X5 + \gamma_s D)\right].l_{toe}^2 \tag{21}$$

$$V_t = \left[1.7\left(\frac{q_{dt}+q_{max}}{2}\right) - 0.9(\gamma_c X5 + \gamma_s D)\right].(l_{toe} - dt) \tag{22}$$

$$M_h = \left[\left(\frac{1.7q+1.4\gamma_c X5+1.4\gamma_s H}{2}\right) + \frac{1.4W_{bs}}{3} - \left(\frac{q_1+2q_{min}}{6}\right)\right].l_{heel}^2 \tag{23}$$

$$V_h = \left[1.7q + 1.4\gamma_c X5 + 1.4\gamma_s H + 1.4\frac{W_{bs}+W_{bsdh}}{2} - 0.9\frac{q_{dh}+q_{min}}{2}\right].(l_{heel} - dh) \tag{24}$$

where $q_2$ is the soil pressure intensity at the junction of stem with toe slab; $q_{max}$ is the maximum soil pressure intensity; $\gamma_c$ is the concrete unit weight; $\gamma_s$ is the soil unit weight; $D$ is the depth of soil in front of wall; $l_{toe}$ is the length of toe slab; $q_{dt}$ is the soil pressure intensity at a distance $dt$ from the junction of the stem and the toe slab; $q$ is the surcharge load; $W_{bs}$ is the maximum resulting load of backfill soil weight; $q_1$ is the soil pressure intensity at the junction of the stem and heel slab; $l_{heel}$ is the length of the heel slab; $W_{bsdh}$ is the resulting load of backfill soil weight at the distance $dh$ from the junction of the stem and heel slab; $q_{dh}$ is the soil pressure intensity at a distance of $dh$; and $q_{min}$ is the minimum soil pressure intensity. The values of $dt$ and $dh$ are computing using (25):

$$dt = dh = X_5 - CC \tag{25}$$

where $CC$ is the depth of concrete cover. The utilized inequality constraints studied by Camp and Akin [55] are presented in Table 1.

Table 1: Inequality constraints

| Constraint | Function |
|---|---|
| $g_1(x)$ | $\frac{FS_{Odesign}}{FS_O} - 1 \leq 0$ |
| $g_2(x)$ | $\frac{FS_{Sdesign}}{FS_S} - 1 \leq 0$ |
| $g_3(x)$ | $\frac{FS_{Bdesign}}{FS_B} - 1 \leq 0$ |
| $g_4(x)$ | $q_{min} \geq 0$ |
| $g_{[5-8]}(x)$ | $\frac{M_u}{M_n} - 1 \leq 0$ |
| $g_{[9-12]}(x)$ | $\frac{V_u}{V_n} - 1 \leq 0$ |
| $g_{[13-16]}(x)$ | $\frac{A_{smin}}{A_s} - 1 \leq 0$ |
| $g_{[17-20]}(x)$ | $\frac{A_s}{A_{smax}} - 1 \leq 0$ |
| $g_{21}(x)$ | $\frac{X_2 + X_3}{X_1} - 1 \leq 0$ |

| $g_{22}(x)$ | $\dfrac{X_6 + X_7}{X_1} - 1 \leq 0$ |
| $g_{23}(x)$ | $\dfrac{l_{db\ stem}}{X_5 - cover} - 1 \leq 0$ or $\dfrac{l_{dh\ stem}}{X_5 - cover} - 1 \leq 0$ |
| $g_{24}(x)$ | $\dfrac{l_{db\ toe}}{X_1 - X_2 - cover} - 1 \leq 0$ or $\dfrac{12 d_{b\ toe}}{X_5 - cover} - 1 \leq 0$ |
| $g_{25}(x)$ | $\dfrac{l_{db\ heel}}{X_2 + X_3 - cover} - 1 \leq 0$ or $\dfrac{12 d_{b\ heel}}{X_5 - cover} - 1 \leq 0$ |
| $g_{26}(x)$ | $\dfrac{l_{db\ key}}{X_5 - cover} - 1 \leq 0$ or $\dfrac{l_{dh\ key}}{X_5 - cover} - 1 \leq 0$ |

For the third phase of the wall design, the geometrical constraints that form a permissible solution space are defined to produce a practical design. The utilized boundary constraints for each example are tabulated according to Table 2:

Table 2: Design variable boundary limitations (from [4]) for numerical simulations

| Design variable | Unit | Example 1 | | Example 2 | |
|---|---|---|---|---|---|
| | | Lower Bound | Upper Bound | Lower Bound | Upper Bound |
| X1 | m | 1.31 | 3.50 | 2.60 | 5.50 |
| X2 | m | 0.44 | 0.78 | 0.87 | 1.56 |
| X3 | m | 0.20 | 0.33 | 0.30 | 0.67 |
| X4 | m | 0.20 | 0.33 | 0.30 | 0.67 |
| X5 | m | 0.27 | 0.33 | 0.54 | 0.67 |
| X6 | m | 1.31 | 3.50 | 2.60 | 5.50 |
| X7 | m | 0.20 | 0.33 | 0.30 | 0.67 |
| X8 | m | 0.20 | 0.33 | 0.30 | 0.67 |
| R1 | - | 1 | 223 | 1 | 223 |
| R2 | - | 1 | 223 | 1 | 223 |
| R3 | - | 1 | 223 | 1 | 223 |
| R4 | - | 1 | 223 | 1 | 223 |

Table 3 shows discrete variables *R1* to *R4*, each referring to the number, *n*, and diameter, *d*, of rebars:

Table 3: Steel reinforcement discrete variables (from [4])

| R1 to R4 | Reinforcement | | | Bars cross-sectional area (cm²) in ascending order |
|---|---|---|---|---|
| | n φ d | quantity | size d (mm) | |
| 1 | 3φ10 | 3 | 10 | 2.356 |
| 2 | 4φ10 | 4 | 10 | 3.141 |
| 3 | 3φ12 | 3 | 12 | 3.392 |
| 4 | 5φ10 | 5 | 10 | 3.926 |
| 5 | 4φ12 | 4 | 12 | 4.523 |
| ⋮ | ⋮ | ⋮ | ⋮ | ⋮ |
| 221 | 16φ30 | 16 | 30 | 113.097 |
| 222 | 17φ30 | 17 | 30 | 120.165 |
| 223 | 18φ30 | 18 | 30 | 127.234 |

## 3 Objective functions

The RCC wall design is defined based on the minimum cost, weight, and CO2 emissions in Eqs. (26)-(28), respectively. Referring to Saribas and Erbatur [4], cost is determined as the estimated cost of steel reinforcement and concrete (including the price of materials and labor), and weight is measured considering an assumed concrete density of 23.5 kN/m$^3$ and a steel density of 78.5kN/m$^3$. CO2 emission is calculated according to the unit concrete emission of 224.94 kg/m$^3$ and steel emission of 2.82 kg/kg, taken from the BEDEC PR/PCT ITeC database, which is available online based on the 2011 Environmental Impact Assessment of Buildings (Barcelona, Spain) [63]. These emissions are multiplied by the weight of steel and the volume of concrete in accordance with the geometry and reinforcement setup [64].

$$Cost = C_s W_s + C_c V_c \tag{26}$$

$$Weight = W_s + 100\gamma_c V_c \tag{27}$$

$$CO_2 = e_s W_s + e_c V_c \tag{28}$$

where $C_s$ and $C_c$ are unit cost of steel and concrete, respectively; $W_s$ is the weight of the reinforcing steel; $V_c$ is the volume of concrete; and $\gamma_c$ is the unit weight of concrete scaled by a factor of 100. The amount of carbon dioxide emissions resulting from RCC wall materials is computed by CO2 as expressed in (24). $e_s$ and $e_c$, taken from the BEDEC PR/PCT ITeC database [63], refer to the unit CO2 emission of reinforcing steel and concrete, respectively.

## 4 Research gaps

Parameter tuning is a common investigation technique used in many previous studies [2] to optimize the computational efforts and better address the RCC wall optimization problem landscape. As discussed, the previous research mostly focused on the performance of an existing evolutionary, swarm intelligence, or PS-inspired approach to the economical design of retaining walls with or without pseudo-static analysis. Based on a review of the relevant studies, further research is warranted to develop a novel adaptive fuzzy based metaheuristic for retaining wall design and to examine the environmental influence of concrete and reinforcing steel and the impact of variation of earthquake coefficients on the optimum design properties of RCC walls.

This study is the first of its kind to consider both the algorithmic developments and pseudo-static analysis for the sustainable, economical, lightweight design of RCC walls, assuming twelve design variables. Our algorithm benefits from novel exploration, exploitation, information sharing strategies, FLSs incorporated in both the parameter adaptation and operator selection procedures, and the adaptive boundary limit strategy. Our proposed method does not require parameter tuning analysis and is inspired by promising results achieved by a similar fuzzy logic-based parameter adaptation and operator selection strategy for solving high-dimensional optimization benchmarks [53]. It is expected to solve the given constrained twelve-dimensional global optimization problem more precisely than the other metaheuristics in the literature.

## 5 Fuzzy set theory

Most metaheuristics require an additional experimental study for parameter tuning [13, 14]. Exploration, exploitation, and information sharing, defined as capabilities of metaheuristics, can cause pre-mature convergence without being controlled by their parameters [15]. Fuzzy Sets (FSs) [65], [66] have been proposed to describe uncertain and vague concepts [67] that originate from noisy decision variables, stochastic search, noisy data, changing environments, among other factors. The idea of type-2 FS was first proposed by Zadeh in 1975 [68], where more than one single membership function and three main components exist: 1) a fuzzifier that receives crisp inputs and produces fuzzy sets, 2) Rule-based Mamdani Inference [69] that defines fuzzy rules, and 3) a defuzzifier that maps from fuzzy sets to crisp outputs. Mendel and Karnik expanded upon this idea and proposed the interval type-2 Mamdani rule-based Fuzzy Logic Systems (FLSs). The type-2 FLSs map to a range, called the Footprint of Uncertainty (FOU), compared to type-1 systems that map to a single value with the lack of uncertainty, and have become the most popular type of FLSs [67].

The hybridization of fuzzy logic systems [68] with search policies in metaheuristics has been effective in OPs, initiating innovative developments in three areas: 1) evolutionary search for the global best fuzzy rule structure or architecture using a metaheuristic [70, 71]; 2) dynamic parameter adaptation in a metaheuristic method using fuzzy rules [72-77]; and 3) dynamic operator selection in a metaheuristic using fuzzy rules [78]. In this study, our proposal relates to (2) and (3), which recently achieved better performance when solving high-dimensional optimization benchmarks despite using fewer computational resources [53].

## 6 Proposed adaptive fuzzy based metaheuristic algorithm

Herein, a novel adaptive fuzzy based metaheuristic is introduced that devises new search components called global learning-based velocity adaptation, enhanced differential evolution-based local search, and universal diversity-based velocity divergence. These policies strengthen three main search capabilities: exploitation, information sharing, and exploration, respectively, and make the approach unique among the other optimization algorithms. Furthermore, fuzzy adaptive parameter adaptation and operation selection are incorporated in each component to guide its search mechanism. To enhance the robustness and efficiency of optimum sustainable-seismic-economical RCC designs, fuzzy adaptation of the velocity equation as well as the colony and imperialist position updating equation in each search policy and fuzzy selection of the policy can be used. The proposed FLSs seek to find appropriate values for control parameters and determine the importance of each search policy's role, particularly helping to balance between the search capabilities, which is necessary to achieve the best possible performance of an optimization algorithm. This method is thus termed Fuzzy Adaptive Global Learning Local Search Universal Diversity (FAGLSUD).

FAGLSUD has some features in common with traditional EAs, such as DE, and SI-based methodologies, such as ICA and PSO. This increases the applicability of the algorithm to many global optimization problems suffering from complex issues, including high-dimensionality, constrained search domain, and discrete-continuous design variables. Like DE where differences between solutions affect a particular solution, FAGLSUD's solutions mimic the effect of gradient descent toward optima in each local space, which is called an empire. Unlike ICA where empires tend to dominate each other until a solo empire remains at the end of convergence, FAGLSUD's empires survive forever, and their characteristics are modified using search policies during the optimization process. Unlike DE that does not necessarily have any built-in tendency to make sub-populations, ICA and FAGLSUD divide the search space into empires. Similar to PSO, FAGLSUD represents each solution as a point in space as a position vector that moves over time and thus, also has a velocity vector. Our empirical study reveals that absence of such a velocity vector and direct update of position vectors may cause poor results; therefore, the velocity vectors are altered then the position vectors are updated. The flowchart of the proposed method is shown in Fig. 3.

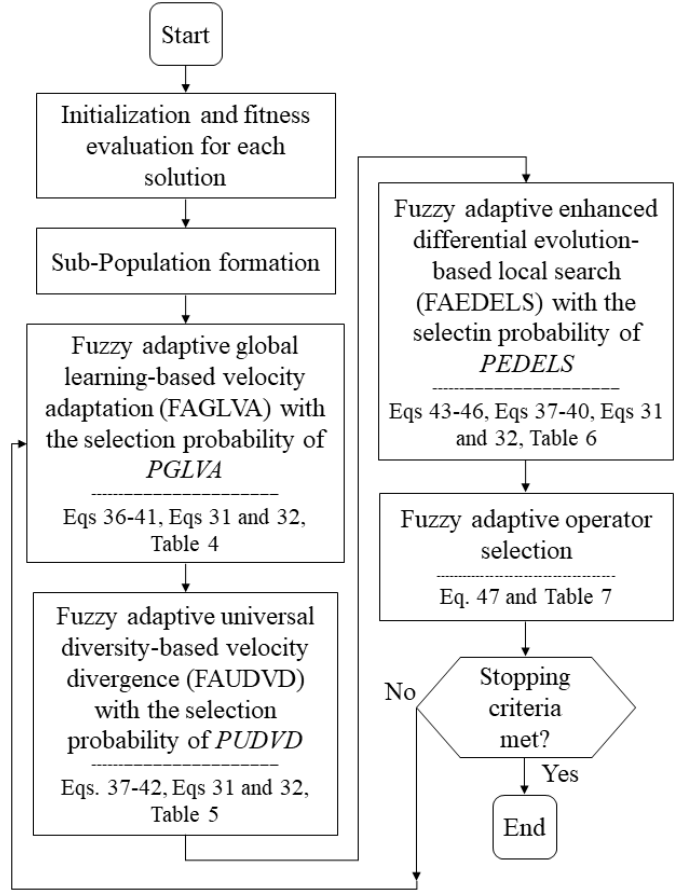

Fig. 3: A flowchart of the proposed FAGLSUD

## 6.1 Initialization

Each solution has position and velocity vector attributes:

$$\begin{cases} P_{Sol,i}^{t=1}{}^d = Var_{min}{}^d + rand(0,1) \times (Var_{max}{}^d - Var_{min}{}^d) \\ V_{Sol,i}^{t=1}{}^d = 0 \end{cases} \quad ; d \in \{1, \ldots, nVar\}, 1 \leq i \leq N_{Pop} \quad (29)$$

where $P_{Sol,i}^{t=1}{}^d$ is the position component of the solution vector $i$ with the dimension of $d$ at iteration $t$, which is 1 for initialization. The restrictions on the boundary of each position vector are determined by $Var_{max}{}^d$ and $Var_{min}{}^d$. The size of each solution vector is determined by $nVar$, which is set to 12 in our study. The total number of initial solutions is set to a fixed value $N_{Pop}$ through the optimization process. The random function $rand(0,1)$ generates random values with uniform distribution within the boundaries assigned to each component.

For each RCC design, the design variables of the $i-th$ position vector $\overline{P_{Sol,i}^t} = [X_1, \ldots, X_8, R_1, \ldots, R_4]$ include twelve components ($X_1, \ldots, X_8$ to describe geometry of the wall, and $R_1, \ldots, R_4$ relating to the steel reinforcement) that follow the geometrical constraints:

$$Example1: \begin{cases} \overline{Var_{max}} = [3.5, 0.78, 0.33, 0.33, 0.33, 4.5, 0.33, 0.33, 223, 223, 223, 223] \\ \overline{Var_{min}} = [1.31, 0.44, 0.2, 0.2, 0.27, 1.31, 0.2, 0.2, 1, 1, 1, 1] \end{cases}$$
$$Example2: \begin{cases} \overline{Var_{max}} = [5.5, 1.56, 0.67, 0.67, 0.67, 5.5, 0.67, 0.67, 223, 223, 223, 223] \\ \overline{Var_{min}} = [2.6, 0.87, 0.3, 0.3, 0.54, 2.6, 0.3, 0.3, 1, 1, 1, 1] \end{cases}$$
(30)

FAGLSUD initializes the position component of each solution vector based on the restrictions defined on the boundary of each element. One may add feasible solutions to the initial population and evaluate its sensitivity to the feasibility of initial candidate RCC retaining wall designs; see [79].

The velocity vector of each solution is initially set to zero.

## 6.2 Fitness evaluation

Each individual evolved in the search space is evaluated by using the fitness function:

$$Fitness_{Sol,i}^t \triangleq Power_{Sol,i}^t = \frac{1}{Penalized\ cost\ function(\overline{P_{Sol,i}})} \tag{31}$$

In particular, the higher the power $Power_{Sol,i}^t$ is, the higher the evaluation of the individual and its corresponding position are. In order to handle the geotechnical and structural design constraints of RCC walls, a penalty term is enforced on the cost function in the denominator:

$$Penalized\ cost\ function(\overline{P_{Sol,i}}) = \begin{cases} cost\ function(\overline{P_{Sol,i}}) + \sum_{j=1}^{N} \lambda \times g^2(j) & ; if\ g(j) > 0 \\ cost\ function(\overline{P_{Sol,i}}) & ; if\ g(j) \leq 0 \end{cases} \tag{32}$$

where $g(j)$ relates to each geotechnical and structural constraint defined for the wall design; and $N$ is the total number of constraints. In our study, RCC designs must deal with 12 design variables and 26 constraints, similar to our previous effort [49]. Therefore, we set the same penalty term $\lambda = 10^{15}$ to perform a fair comparison with BBO's low-cost and low-weight results [49]. For a lower number of design variables and nonlinear constraints involved in the design of RCC, lower penalty factor might be more appropriate. For instance, when solving a six-dimensional RCC design problem with five constraints, a lower penalty term was added to calculations of the low-Cost design results [2].

## 6.3 Sub-population formation

The sub-population (i.e., empire) formulation helps to provide better interaction and information sharing between solutions (i.e., countries). The $N_{Imp}$ best countries in terms of power build empires and the other countries will be $N_{Col}$ colonies [17]. Unlike ICA, these numbers are kept the same through the optimization process.

$$N_{Col} = N_{Pop} - N_{Imp} \tag{33}$$

Each imperialist starts taking control of other countries (called colonies) and forms the initial empires [80].

$$Normalized\ Power_{Imp,j} = \max_{j} Power_{Imp,j} - Power_{Imp,j} \tag{34}$$

where the power of $j-th$ imperialist $Power_{Imp,j}$ is normalized by using the finest imperialist $\max_{j} Power_{Imp,j}$. The main colonies are separated into empires based on normalized power for the $k-th$ empire. The $k-th$ imperialist is selected using a *Roulette Wheel Selection* procedure to attract $N_k$ number of colonies and form the $k-th$ empire:

$$\begin{cases} k = Roulette\ Wheel\ Selection\left(\dfrac{Normalized\ Power_{Imp,j}}{\sum_{j=1}^{N_{Imp}}(Normalized\ Power_{Imp,j})}\right) \\ N_k = Round\left(\left|\dfrac{Normalized\ Power_{Imp,j}}{\sum_{j=1}^{N_{Imp}}(Normalized\ Power_{Imp,j})}\right| \times N_{Col}\right) \end{cases} \quad (35)$$

### 6.4 Fuzzy adaptive global learning-based velocity adaptation

Inspired by the position and velocity update process devised in the PSO and assimilation operator in the ICA, we propose a new operator called global learning-based velocity adaptation (GLVA). GLVA contributes to exploitation (intensification) of solutions in search space. As shown in Fig. 4 and represented by Eq. (36), GLVA reflects the movements of: 1) each colony toward its best-known experience and its associated imperialist, and 2) each imperialist toward its best-known experience and the global best imperialist. The velocity components of colony $i$ and imperialist $j$ denoted as $\overrightarrow{V_{Col,i}^{t+1}}$ and $\overrightarrow{V_{Imp,J}^{t+1}}$ are updated through the following learning process:

$$\begin{cases} \overrightarrow{V_{Col,i}^{t+1}} = \beta_1 r_1\left(\overrightarrow{P_{Imp,J}^t} - \overrightarrow{P_{Col,i}^t}\right) + c_1 r_2\left(\overrightarrow{P_{lbestCol,i}^t} - \overrightarrow{P_{Col,i}^t}\right) \\ \overrightarrow{V_{Imp,J}^{t+1}} = \beta_2 r_1\left(\overrightarrow{P_{Globalbest}^t} - \overrightarrow{P_{Imp,J}^t}\right) + c_2 r_2\left(\overrightarrow{P_{lbestImp,J}^t} - \overrightarrow{P_{Imp,J}^t}\right) \end{cases}, if\ r \leq PFAGLVA \quad (36)$$

where each colony or imperialist learns from its best-known experience $\overrightarrow{P_{lbestCol,i}^t}$ and $\overrightarrow{P_{lbestImp,J}^t}$; $r_1$ and $r_2$ represent a random number with uniform distribution between 0 and 1 that make these movements stochastic; $c_1$ and $c_2$ focus on cognitive personal learning; and $\beta_1$ and $\beta_2$ accelerate the social learning aspect. The best-ever-known solution, namely the global best imperialist, is located at the position of $\overrightarrow{P_{Globalbest}^t}$, which helps to guide the search for each imperialist. The probability to select the GLVA operator is set by $PFAGLVA$, which will be explained in the last phase.

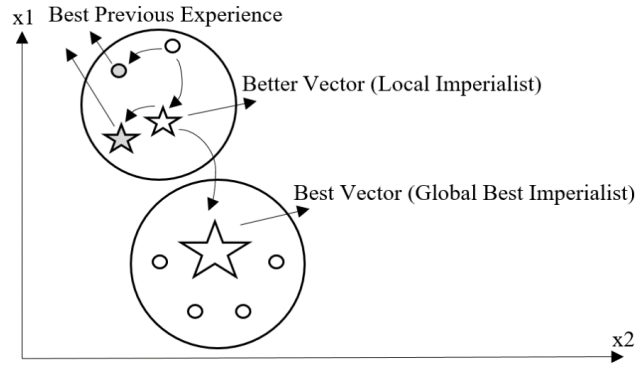

Fig. 4: Representation of GLVA in a sample 2-dimmensional search space

### 6.4.1 Adaptive velocity limit function and position update

An interesting study by Azad, S.K. and O. Hasançebi [81] formulated an upper bound strategy, which considers the current best design as the upper bound for the forthcoming candidates in order to eliminate unnecessary structural analysis and associated fitness computations for those candidates that have no chance of surpassing the best solution. Inspired by this method, we impose the global best solution in an adaptive velocity limit function (AVLF) to reduce the total number of RCC analyses by avoiding unnecessary analyses during RCC optimization. The AVLF provides high exploration, and thus prevents the countries from getting trapped in local minima during the early stage of the optimization process; it also provides high exploitation to improve the overall convergence behavior at the later stage of convergence.

$$\begin{cases} \overline{Vel_{max}} = +\alpha \left(\frac{\overline{Var_{max}} - \overline{Var_{min}}}{\overline{Var_{max}}}\right) \left|\frac{\overline{P_{Globalbest}^{t}} - \overline{P_{Sol,i}^{t}}}{t}\right|, \forall\, \overline{P_{Sol,i}^{t}} \neq \overline{P_{Globalbest}^{t}} \\ \overline{Vel_{max}} = 0, \forall\, \overline{P_{Sol,i}^{t}} = \overline{P_{Globalbest}^{t}} \\ \overline{Vel_{min}} = -\overline{Vel_{max}} \\ \overline{Vel_{min}} < \overline{V_{Sol,i}^{t}} = [V_1, \dots, V_{nVar}] < \overline{Vel_{max}} \end{cases} \quad (37)$$

Based on the distance of a solution at iteration $t$ from that at iteration $t + 1$, a velocity is assigned that changes the solution's position to bring it closer to the global best solution. The magnitude and direction of the velocity vector can be suitably changed by AVLF, resulting in the solution moving toward the optimal solution with faster convergence. The higher upper bound increases the exploratory capabilities, while the lower upper bound increases the exploitation of the solutions in the search space. High exploration decreases the likelihood of the search ending up at a local minimum, while increasing the convergence time. Conversely, high exploitation significantly reduces convergence time, while it reduces the global search capability of the algorithm and increases the likelihood of the search ending up at a local minimum. Reducing the convergence time becomes more important for solving large-scale structural optimization problems consisting of many design variables and/or large discrete sets [82]. The $\overline{V_{Sol,i}^{t}}$ for the $i-$ th solution can be adaptively clamped by specifying the upper and lower bounds maintained as above. Eq. (37) shows the velocity limits that are linearly decreased depending on the number of iterations, expecting to meet the requirements for high exploration during the earlier part of the convergence history and high exploitation towards the end of convergence history. It gives the current search location $\overline{P_{Sol}^{t}}$ a tendency to keep moving in the direction of the global best solution, $\overline{P_{Globalbest}^{t}}$ using fewer computational resources. The $\alpha$ was set to 10 based on empirical study of several values.

Since the evolved solutions may lie outside the search-space boundary, Eq. (38) is applied:

$$\begin{cases} \begin{cases} V_{Col,i}^{t+1\,d} = Vel_{max}^{\,d} & if\ V_{Col,i}^{t+1\,d} > Vel_{max}^{\,d} \\ V_{Col,i}^{t+1\,d} = Vel_{min}^{\,d} & if\ V_{Col,i}^{t+1\,d} < Vel_{min}^{\,d} \end{cases} \\ \begin{cases} V_{Imp,j}^{t+1\,d} = Vel_{max}^{\,d} & if\ V_{Imp,j}^{t+1\,d} > Vel_{max}^{\,d} \\ V_{Imp,j}^{t+1\,d} = Vel_{min}^{\,d} & if\ V_{Imp,j}^{t+1\,d} < Vel_{min}^{\,d} \end{cases} \end{cases} \quad (38)$$

Then, each position is updated by Eq. (39):

$$\begin{cases} \overline{P_{Col,i}^{t+1}} = \overline{P_{Col,i}^{t}} + \overline{V_{Col,i}^{t+1}} \\ \overline{P_{Imp,j}^{t+1}} = \overline{P_{Imp,j}^{t}} + \overline{V_{Imp,j}^{t+1}} \end{cases} \quad (39)$$

If the new position component $\overline{P_{Col,i}^{t+1}}$ or $\overline{P_{Imp,j}^{t+1}}$ at the iteration $t+1$ crosses the border (i.e., $(Var_{min}, Var_{min})$), their corresponding velocity component, $\overline{V_{Col,i}^{t+1}}$ or $\overline{V_{Imp,j}^{t+1}}$, will be reversed:

$$\begin{cases} \begin{cases} P_{Col,i}^{t+1\ d} = Var_{max}^{\ d}, V_{Col,i}^{t+1\ d} = -V_{Col,i}^{t+1\ d} & if\ P_{Col,i}^{t+1\ d} > Var_{max}^{\ d} \\ P_{Col,i}^{t+1\ d} = Var_{min}^{\ d}, V_{Col,i}^{t+1\ d} = -V_{Col,i}^{t+1\ d} & if\ P_{Col,i}^{t+1\ d} < Var_{min}^{\ d} \end{cases} \\ \begin{cases} P_{Imp,j}^{t+1\ d} = Var_{max}^{\ d}, V_{Imp,j}^{t+1\ d} = -V_{Imp,j}^{t+1\ d} & if\ P_{Imp,j}^{t+1\ d} > Var_{max}^{\ d} \\ P_{Imp,j}^{t+1\ d} = Var_{min}^{\ d}, V_{Imp,j}^{t+1\ d} = -V_{Imp,j}^{t+1\ d} & if\ P_{Imp,j}^{t\ d} < Var_{min}^{\ d} \end{cases} \end{cases} \quad (40)$$

**6.4.2 Parameter adaptation by fuzzy adaptive mechanism**

Having established the GLVA search policy, the proposal is to adapt the parameters using FLSs; namely Fuzzy Adaptive GLVA (FAGLVA). The mechanism starts with Membership Functions (MFs) that are depicted in Fig. 5. MFs map the input quantities to descriptive words {low, medium, high}, following smooth changes rather than sharp switching between each status. MFs provide an alternative way to quantify the amount of uncertainty instead of using probabilistic measures, and through that, complex optimization problems can be solved in less time [83].

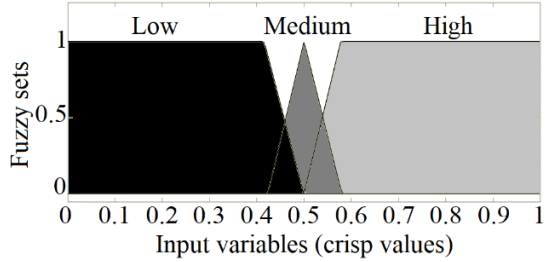

Fig. 5: Three MFs for mapping input quantities into linguistic terms {Low, Medium, and High}

The input quantities, namely *Normalized Relative Powers NRP₁ to NRP₄*, are set in the range between 0 and 1, which helps to screen the fitness functions.

$$\begin{cases} NRP_1(\overline{P_{Col,i}^t}) = \left|\frac{Power(P_{Imp,J}^t) - Power(P_{Col,i}^t)}{Power(P_{Globalbest}^t)}\right| \\ NRP_2(\overline{P_{Col,i}^t}) = \left|\frac{Power(P_{lbestCol,i}^t) - Power(P_{Col,i}^t)}{Power(P_{Globalbest}^t)}\right| \\ NRP_3(\overline{P_{Imp,J}^t}) = \left|\frac{Power(P_{Globalbest}^t) - Power(P_{Imp,J}^t)}{Power(P_{Globalbest}^t)}\right| \\ NRP_4(\overline{P_{Imp,J}^t}) = \left|\frac{Power(P_{lbestImp,J}^t) - Power(P_{Imp,J}^t)}{Power(P_{Globalbest}^t)}\right| \\ Normalized\ Iteration = NIT = \frac{t}{Maximum\ Iteration} \end{cases} \quad (41)$$

Next, a set of fuzzy rules (i.e., if-then statements) is formed for each fuzzy inference system (FIS), which correlates the input functions and output parameters, as shown in Table 4. For instance, rule #1 points out a specific convergence status at the beginning and middle of the convergence time (i.e., *NIT* is low or medium). Specifically, low *NRP₁* means that the current search location (i.e., position of a colony) is close to its relative imperialist; low *NRP₂* indicates that the colony is close to its best-known experience; low $NRP_3$ refers to a close distance between an imperialist and the global best imperialist; and low $NRP_4$ shows that an imperialist is close to its best-known experience. In order to avoid getting stuck in a local optimum solution, a low credence should be given to the current convergence status, thus $\beta_1, \beta_2, c_1$, and $c_2$ get low values, and the overall performance is assumed to improve. As another example, rule #18 indicates the status at the end of convergence, and much focus should be given to the information gained at that time rather than learning new information. Therefore, cognitive personal best coefficients get *high* values, while social learning coefficients get *low* values. This will enhance the convergence speed.

Table 4: If-then statements for adapting GLVA's parameters

| # Rule | If (Status Indicators) | | | | | Then (Parameter Adaptation for the next $t$) | | | |
|---|---|---|---|---|---|---|---|---|---|
| | $NRP_1$ | $NRP_2$ | $NRP_3$ | $NRP_4$ | NIT | $\beta_1$ | $c_1$ | $\beta_2$ | $c_2$ |
| 1 | L | L | L | L | L or M | L | L | L | L |
| 2 | L | L | L | H | L or M | L | L | L | H |
| 3 | L | L | H | L | L or M | L | L | H | L |
| 4 | L | L | H | H | L or M | L | L | H | H |
| 5 | L | H | L | L | L or M | L | H | L | L |
| 6 | L | H | L | H | L or M | L | H | L | H |
| 7 | L | H | H | L | L or M | L | H | H | L |
| 8 | L | H | H | H | L or M | L | H | H | H |
| 9 | H | L | L | L | L or M | H | L | L | L |
| 10 | H | L | L | H | L or M | H | L | L | H |
| 11 | H | L | H | L | L or M | H | L | H | L |
| 12 | H | L | H | H | L or M | H | L | H | H |
| 13 | H | H | L | L | L or M | H | H | L | L |
| 14 | H | H | L | H | L or M | H | H | L | H |
| 15 | H | H | H | L | L or M | H | H | H | L |
| 16 | H | H | H | H | L or M | H | H | H | H |
| 17 | M | M | M | M | L or M | M | M | M | M |
| 18 | Any | Any | Any | Any | H | L | H | L | H |

Finally, the linguistic terms are mapped into the real values. As demonstrated in Fig. 6, the centroid defuzzification method is used to measure the center of the areas under the MF curves.

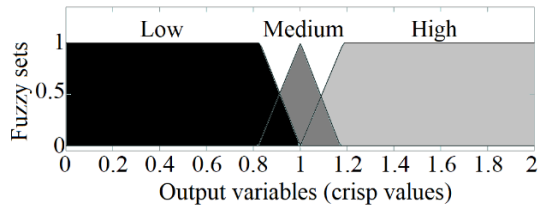

Fig. 6: Three MFs for mapping linguistic terms {Low, Medium, High} from fuzzy sets to crisp values

## 6.5 Fuzzy adaptive universal diversity-based velocity divergence

Motivated by the concept of comprehensive learning originally developed by Liang, Qin et al. [84] and enhanced exploration capability of the standard PSO, we propose a new operator called universal diversity-based velocity divergence (UDVD). UDVD contributes to the exploration (diversification) of solutions in search space since the best position ever found by other colonies and imperialists can potentially be employed as the exemplars to guide the movement direction of a candidate colony, imperialist, and the global best imperialist. The UDVD is proposed to prevent fast convergence of search agents toward local optimum solutions and discourage the method from premature convergence.

As represented by Eq. (42), UDVD reflects three learning mechanisms–the $d-th$ dimension of 1) a colony $P_{Col,i}^{t}{}^{d}$, 2) an imperialist $P_{Imp,j}^{t}{}^{d}$, and 3) the global best imperialist $P_{Globalbest}^{t}{}^{d}$ will learn from other colonies and imperialists. These mechanisms lead to the following movements:

$$\begin{cases} V_{Col,i}^{t+1}{}^{d} = w_1\left\{\left(V_{Col,i}^{t}{}^{d}\right) + r_1\left(P_{lbestCol,l}^{t}{}^{d} - P_{Col,i}^{t}{}^{d}\right)\right\}, \forall l \neq i \\ V_{Imp,j}^{t+1}{}^{d} = w_2\left\{\left(V_{Imp,j}^{t}{}^{d}\right) + r_2\left(P_{lbestImp,k}^{t}{}^{d} - P_{Imp,j}^{t}{}^{d}\right)\right\}, \forall k \neq j \quad , if\ r \leq PFAUDVD \\ V_{Globalbest}^{t+1}{}^{d} = w_3\left\{V_{Globalbest}^{t}{}^{d} + r_3\left(P_{lbestImp,k}^{t}{}^{d} - P_{Globalbest}^{t}{}^{d}\right)\right\} \end{cases} \quad (42)$$

where $w_1, w_2$, and $w_3$ are learning coefficients that help diversification of colonies, imperialists, and the global best imperialist in the search space, respectively. The random parameters $r_1, r_2$, and $r_3$ make the movements stochastic. The probability to select the UDVD operator is set by $PFAUDVD$, which will be described in the last phase. A $Roulette-Wheel-Selection$ procedure is applied to select the $l-th$ colony and $k-th$ imperialist based on their power. Stronger colonies or imperialists are more likely to be selected. The UDVD's AVLF and position update process are similar to those of GLVA.

### 6.5.1 Parameter adaptation by fuzzy adaptive mechanism

Having established the UDVD search policy, the proposal is to adapt the control parameters using FLSs; namely the FAUDVD policy. To this end, as shown in Table 5, a new set of fuzzy rules correlating the input quantities to the parameters is devised.

Table 5: If-then statements for adapting parameters of the FAUDVD operator

| #Rule | If (Status Indicators) | | | | | Then (Parameter Adaptation for the next $t$) | | |
|---|---|---|---|---|---|---|---|---|
| | $NRP_1$ | $NRP_2$ | $NRP_3$ | $NRP_4$ | NIT | $w_1$ | $w_2$ | $w_3$ |
| 1 | L | L | L | L | L or M | L | L | L |
| 2 | L | L | L | H | L or M | L | L | L |
| 3 | L | L | H | L | L or M | L | H | L |
| 4 | L | L | H | H | L or M | L | H | L |
| 5 | L | H | L | L | L or M | L | L | L |
| 6 | L | H | L | H | L or M | L | L | L |
| 7 | L | H | H | L | L or M | L | H | L |
| 8 | L | H | H | H | L or M | L | H | L |
| 9 | H | L | L | L | L or M | H | L | L |
| 10 | H | L | L | H | L or M | H | L | L |
| 11 | H | L | H | L | L or M | H | H | L |
| 12 | H | L | H | H | L or M | H | H | L |
| 13 | H | H | L | L | L or M | H | L | L |
| 14 | H | H | L | H | L or M | H | L | L |
| 15 | H | H | H | L | L or M | H | H | L |
| 16 | H | H | H | H | L or M | H | H | H |
| 17 | M | M | M | M | L or M | M | M | M |
| 18 | Any | Any | Any | Any | H | L | L | L |

For example, rule #1 indicates an ideal convergence status, where a colony is closest to its local imperialist and its best-known experience, and all input quantities $NRP_1$ to $NRP_4$ are low in a sense that all designated distances are low. To keep this status, a low credence must be given to the diversification factors. An imperialist is closest to the global best imperialist and its best-known experience; thus, $w_1, w_2$, and $w_3$ get low values and the overall performance is expected to enhance. Oppositely, rule #16 refers to the worst convergence status where all these distances are high. To relieve the algorithm from getting stuck in local optimum solutions, a high credence must be given to the diversification factors. Thus, $w_1, w_2$, and $w_3$ get high values, and the overall performance is expected to enhance. As another exemplar, rule #18 indicates the end of convergence time, so less focus must be given to the diversification factors. Thus, $w_1, w_2$, and $w_3$ get low values and the convergence rate is expected to improve. The defuzzification step is similar to that of FAGLVA as depicted in Fig. 6.

## 6.6 Fuzzy adaptive enhanced differential evolution-based local search

Inspired by the differential evolution method initially proposed by Storn [24] for optimizing real-valued objective functions at high convergence rate and recently advanced by Hadi, M et al. [85] for solving high-dimensional global optimization problems, our proposal is to model a new operator named enhanced differential evolution based local search (EDELS). The EDELS contributes to the information sharing capability (i.e., a combination of exploration and exploitation capabilities) of the method and, thus, enhances the overall performance since it mimics the effect of gradient descent for movements of solutions toward optima and against poor solutions. EDELS executes three phases: *mutation*, *crossover*, and *selection*. The mutant colony or imperialist vectors $\overline{V_{MutantCol,k}^{t+1}}$ or $\overline{V_{MutantImp,k}^{t+1}}$ are a result of the *mutation* phase in the $k-th$ empire. These vectors are considered local search steps to help diversify solutions in each empire and increase the chance for the main randomly chosen candidate colony or imperialist vector $\overline{P_{Col_{r3,k}}^t}$ or $\overline{P_{Imp_{r3,k}}^t}$ to locate more optimality:

$$\begin{cases} \overline{V_{MutantCol,k}^{t+1}} = F_1 \cdot \left(\overline{P_{Col_{r1,k}}^t} - \overline{P_{Col_{r2,k}}^t}\right) + F_2 \cdot \left(\overline{P_{Imp_k}^t} - \overline{P_{Col_{r3,k}}^t}\right) + F_3 \cdot \left(\overline{P_{Col_{r3,k}}^t} - \overline{P_{wCol,k}^t}\right) \\ \overline{V_{MutantImp,k}^{t+1}} = F_4 \cdot \left(\overline{P_{Imp_{r1,k}}^t} - \overline{P_{Imp_{r2,k}}^t}\right) + F_5 \cdot \left(\overline{P_{Globalbest}^t} - \overline{P_{Imp_{r3,k}}^t}\right) + F_6 \cdot \left(\overline{P_{Imp_{r3,k}}^t} - \overline{P_{wImp}^t}\right), if\ r \leq PFAEDELS \\ \overline{P_{Col_{r1,2,3,k}}^t} \neq \overline{P_{wCol,k}^t},\ \overline{P_{Imp_{r1,2,3,k}}^t} \neq \overline{P_{wImp}^t} \end{cases} \quad (43)$$

where the scaling factors $F_1 \ldots F_6$ act as amplifiers at the difference terms that are adaptively tuned based on fuzzy logic in this study. According to Storn and Price [86], these parameters should be in the range of 0-2. Two other randomly chosen colonies and imperialists are: $\overline{P_{Col_{r1,k}}^t}, \overline{P_{Col_{r2,k}}^t}$ and $\overline{P_{Imp_{r1,k}}^t}$ and $\overline{P_{Imp_{r2,k}}^t}$ that help to provide strong exploration in a sub-region.

Next, in the *crossover* phase, the mutant vector is mixed with the main randomly selected solution, yielding the trial vector:

$$\begin{cases} V_{TrialCol}^{t+1}{}^d = \begin{cases} V_{MutantCol}^{t+1}{}^d, & if\ r \leq PFAEDELS \\ V_{Col_{r3}}^t{}^m, & Otherwise \end{cases} \\ V_{TrialImp}^{t+1}{}^d = \begin{cases} V_{MutantImp}^{t+1}{}^d, & if\ r \leq PFAEDELS \\ V_{Imp_{r3}}^t{}^m, & Otherwise \end{cases} \end{cases} ; d = rand\ integer\ \{1, \ldots, n_x\} \quad (44)$$

Having established the velocity components, the positions are achieved by:

$$\begin{cases} \overline{P_{TrialCol}^{t+1}} = \overline{P_{Col_{r3}}^{t}} + \overline{V_{TrialCol}^{t+1}} \\ \overline{P_{TrialImp}^{t+1}} = \overline{P_{Imp_{r3}}^{t}} + \overline{V_{TrialImp}^{t+1}} \end{cases} \quad (45)$$

Similarly, Eqs. (37)-(40) can be used to guide solutions toward the global best solution and also to prevent the components from lying outside the search-space boundary. Finally, the selection phase is carried out by comparing the power of the main randomly selected colony $\overline{P_{Col_{r3}}^{t}}$ or imperialist $\overline{P_{Imp_{r3}}^{t}}$ with the power of their corresponding trial vectors $\overline{P_{TrialCol}^{t+1}}$ or $\overline{P_{TrialImp}^{t+1}}$ and selecting a stronger colony or imperialist. The final solution vector is called the target vector [24].

**6.6.1 Parameter adaptation by fuzzy adaptive mechanism**

As mentioned earlier, information sharing without any control can cause pre-mature convergence [87]. Herein, our proposal is to adapt the control parameters using FLS; namely the FAEDELS policy. For this purpose, a new group of fuzzy rules correlating a new set of input quantities to the parameters is proposed. Eq. (46) represents the quantities, and Table 6 lists the fuzzy rules:

$$\begin{cases} NRP_5\left(\overline{P_{Col_{r3,k}}^{t}}\right) = \left|\frac{Power\left(\overline{P_{Imp,k}^{t}}\right) - Power\left(\overline{P_{Col_{r3,k}}^{t}}\right)}{Power\left(P_{Globalbest}^{t}\right)}\right| \\ NRP_6\left(\overline{P_{Col_{r3,k}}^{t}}\right) = \left|\frac{Power\left(\overline{P_{Col_{r3,k}}^{t}}\right) - Power\left(\overline{P_{wCol,k}^{t}}\right)}{Power\left(P_{Globalbest}^{t}\right)}\right| \\ NRP_7\left(\overline{P_{Imp_{r3}}^{t}}\right) = \left|\frac{Power\left(P_{Globalbest}^{t}\right) - Power\left(\overline{P_{Imp_{r3}}^{t}}\right)}{Power\left(P_{Globalbest}^{t}\right)}\right| \\ NRP_8\left(\overline{P_{Imp_{r3}}^{t}}\right) = \left|\frac{Power\left(\overline{P_{Imp_{r3}}^{t}}\right) - Power\left(P_{wImp}^{t}\right)}{Power\left(P_{Globalbest}^{t}\right)}\right| \end{cases} \quad (46)$$

where $NRP_5 \dots NRP_8$ respectively indicate the normalized difference between the power of: 1) the main randomly selected colony $\overline{P_{Col_{r3,k}}^{t}}$ of the $k-th$ empire with that of its associated imperialist, 2) the main randomly selected colony with that of the worst colony $\overline{P_{wCol,k}^{t}}$ in the population, 3) the main randomly chosen imperialist $\overline{P_{Imp_{r3}}^{t}}$ with that of the global best imperialist $\overline{P_{Globalbest}^{t}}$, and 4) the main randomly chosen imperialist with that of the worst imperialist.

Table 6: If-then statements for adapting parameters of the FAEDELS policy

| #Rule | If (Status Indicators) | | | | | Then (Parameter Adaptation for the next *t*) | | | | | |
|---|---|---|---|---|---|---|---|---|---|---|---|
| | $NRP_5$ | $NRP_6$ | $NRP_7$ | $NRP_8$ | NIT | $F_1$ | $F_2$ | $F_3$ | $F_4$ | $F_5$ | $F_6$ |
| 1 | L | L | L | L | L or M | H | H | L | H | H | L |
| 2 | L | L | L | H | L or M | H | H | L | L | H | L |
| 3 | L | L | H | L | L or M | H | H | L | H | L | L |
| 4 | L | L | H | H | L or M | H | H | L | H | L | H |
| 5 | L | H | L | L | L or M | L | H | H | H | H | L |
| 6 | L | H | L | H | L or M | L | H | H | L | H | H |
| 7 | L | H | H | L | L or M | L | H | H | H | L | L |
| 8 | L | H | H | H | L or M | L | H | H | H | L | H |
| 9 | H | L | L | L | L or M | H | L | L | H | H | L |

| 10 | H | L | L | H | L or M | H | L | L | L | H | H |
| 11 | H | L | H | L | L or M | L | H | H | L | H | H |
| 12 | H | L | H | H | L or M | H | L | L | H | L | H |
| 13 | H | H | L | L | L or M | H | L | H | H | H | L |
| 14 | H | H | L | H | L or M | H | L | H | L | H | H |
| 15 | H | H | H | L | L or M | H | L | H | H | L | L |
| 16 | H | H | H | H | L or M | H | L | H | H | L | H |
| 17 | M | M | M | M | L or M | M | M | M | M | M | M |
| 18 | Any | Any | Any | Any | H | L | H | L | L | H | L |

For instance, rule #11 points to a poor convergence status at the earlier or middle time of convergence (i.e., *NIT* is low or medium), whereby $NRP_5$ is high, $NRP_6$ is low, $NRP_7$ is high, and $NRP_8$ is low. Consequently, their related parameters vary in a way that helps the algorithm escape from local optima; for example, a high credence is given to the distance between a randomly selected colony and its associated imperialist. As a result, the overall performance has improved.

### 6.7 Fuzzy adaptive operator selection

Employing all search policies for the evolution of a large number of search agents causes low-speed convergence [88]. Our proposal, in this part, is to present a new list of fuzzy rules, as shown in Table 7. A quantity denoted as *stagnation* computed by Eq. (47) measures the dynamics status of the global best imperialist $\overline{P_{Globalbest}^t}$ during each *time window(tw)*, which means that when the time t is a factor of the time interval tw, the *stagnation* is calculated. A high *stagnation* value at the early or middle iteration time implies the low usefulness of utilized search policy over the past $tw$. Therefore, the fuzzy rules are implemented such that the probability of selection of each search policy is reversed, and the overall performance for the next $tw$ is expected to enhance.

$$\begin{cases} Stagnation(\overline{P_{Globalbest}^{tw}}) = 1 - \frac{\underset{tw}{Max}\, Power(\overline{P_{Globalbest}^t}) - \underset{tw}{Min}\, Power(\overline{P_{Globalbest}^t})}{\underset{tw}{Max}\, Power(\overline{P_{Globalbest}^t})} \\ t = \{1 \leq t \leq max\ iteration\ |\ remainder\ is\ 0\ when\ dividing\ t\ by\ tw\} \end{cases} \quad (47)$$

where the normalized difference between the strongest and weakest global imperialist within $tw$ is calculated using $\underset{tw}{max}$ and $\underset{tw}{min}\, Power(\overline{P_{Globalbest}^t})$ terms, respectively. Based on our empirical study, the $tw$ is set as ten for calculating the stagnation on t = 10, 20, 30, ..., maximum iteration.

Table 7: If-then statements for selecting operators of the algorithm

| | If (within the past $tw$) | | | | | Then (for the next $tw$) | | |
|---|---|---|---|---|---|---|---|---|
| # Rule | NIT | Stagnation | PFAGLVA | PFAUDVD | PFAEDELS | PFAGLVA | PFAUDVD | PFAEDELS |
| 1 | L | L | L | L | L | L | L | L |
| 2 | | | L | L | H | L | L | H |
| 3 | | | L | H | L | L | H | L |
| 4 | | | L | H | H | L | H | H |
| 5 | | | H | L | L | H | L | L |
| 6 | | | H | L | H | H | L | H |
| 7 | | | H | H | L | H | H | L |
| 8 | | | H | H | H | H | H | H |
| 9 | | H | L | L | L | H | H | H |
| 10 | | | L | L | H | H | H | L |
| 11 | | | L | H | L | H | L | H |
| 12 | | | L | H | H | H | L | L |
| 13 | | | H | L | L | L | H | H |
| 14 | | | H | L | H | L | H | L |
| 15 | | | H | H | L | L | L | H |
| 16 | | | H | H | H | L | L | L |
| 17 | M | M | M | M | M | M | M | M |
| 18 | H | L | L | L | L | H | H | H |
| 19 | | | L | L | H | H | H | L |
| 20 | | | L | H | L | H | L | H |
| 21 | | | L | H | H | H | L | L |
| 22 | | | H | L | L | L | H | H |
| 23 | | | H | L | H | L | H | L |
| 24 | | | H | H | L | L | L | H |
| 25 | | | H | H | H | L | L | L |
| 26 | | H | L | L | L | L | L | L |
| 27 | | | L | L | H | L | L | H |
| 28 | | | L | H | L | L | H | L |
| 29 | | | L | H | H | L | H | H |
| 30 | | | H | L | L | H | L | L |
| 31 | | | H | L | H | H | L | H |
| 32 | | | H | H | L | H | H | L |
| 33 | | | H | H | H | H | H | H |

According to Table 7, rules #1 to #8 show a low *stagnation* value for the global best imperialist found within a *tw* at the beginning of executions, indicating the high usefulness of the search policies ever executed. Thus, the probability of implementing each operator should be kept the same. Rules #18 to #25 refer to a low stagnation value for the global best imperialist within a *tw* at the end of executions. In that case, more emphasis should be given on intensification rather than diversification, thus, the probabilities of applying search policies must be reversed. For rules #9 to #16, the stagnation is high at early stages of convergence, so all probabilities of applying search policies should be reversed. Finally, for rules #26 to #33, the stagnation is high, but that is at the end of convergence time, so the probabilities of using search policies are kept the same as before.

## 7    Simulation and numerical studies

In this section, we report on the efficacy of the proposed algorithm using two numerical RCC wall design optimization examples. Each was conducted each under nine pseudo-static loading cases simulated in MATLAB R2019b and implemented on Intel ® Core™ i5-7300U CPU @ 2.60GHz, 8.00 GB RAM. The classical and recently proposed EA, SI, and PS exemplars, which showed good performance in this optimization problem landscape, were also investigated for comparison. All reported results are from a series of 101 independent runs, including mean results along with the standard deviation (SD). Since metaheuristics follow a stochastic nature, it is impossible to determine their efficiency by a single simulation run [30]. Further, and more importantly, the performance can be significantly improved by using feasible solutions, and by increasing the number of repetitions (re-starts). For all methods, the population size and number of generations (or iterations as the stopping criterion for each run) were set equal to 50 and 1000, respectively. Repeating the algorithms 1000 times will ensure enough accuracy for most optimization applications. However, comparison of metaheuristics by using the best optimization results of 101 independent runs may be within the limits of numerical errors resulting from the accuracy of the calculations, including rounding of decimal numbers. Consequently, we compare the optimization methods using the mean of 101 independent runs, rather than the best solution. It should be noted that these settings are in accordance with our previous low-cost and low-weight study on 12-dimensional RCC design [49] to ensure a fair comparison.

The metaheuristics were calibrated in prior studies; further details are provided in Table 8. For instance, the initial temperature parameter of modified SA was adjusted to 1/20 the cost of the initial best solution in the RCC design study conducted by Yepes, V. et al. [89], which was inspired by the adjustment method devised by Medina [90]. Therefore, if the acceptance percentage of higher energy solutions is determined within 20% to 40%, the initial temperature is doubled, otherwise the initial temperature is halved. Kaveh, A. et al. investigated the dynamic adaptation process of improved HS (IHS) and obtained better results in RCC wall design by changing two parameters, the pitch adjusting rate and bandwidth, dynamically over the iteration number [35]. In this regard, FAGLSUD's parameters were adapted based on fuzzy logic theory, and thus, our method is needless of parameter tuning analysis.

Table 8: The control parameters of the metaheuristics utilized in this study, as well as common settings including population size nPop = 50, maximum number of iterations = 1000, and the number of design variables nVar = 12

| Algorithms | Parameters |
|---|---|
| FPA[*][2] | Power law index ($\lambda$) = 1.5, Probability value (p) = 0.5, Light absorption coefficient Gamma $\gamma$ = 0.5 |
| GWO[*][41] | Number of grey wolves = 50 |
| LFBBO[*][38] | Light absorption coefficients; Gamma 1 ($\Gamma_1$) = 1.329340388179137, Gamma 2 ($\Gamma_2$) = 0.906402477055477, Stability (Levy) index ($\beta$) (Attraction coefficient base value) = 1.5, Number of new solutions = 40 (#new Habitats), Number of kept solutions (habitats) = 10, Mutation rate (m) = 0.1 |
| BBO [49] | Number of new solutions = 40 (#new Habitats), Number of kept solutions (habitats) = 10, Mutation rate (m) = 0.1 |
| DE [47] | Scaling factor ($F$) = 0.2, Crossover probability ($pCR$) = 0.01 |
| ICA[*][17] | Number of sub-populations = 20, Assimilation factor ($\beta$) = 2.1, Revolution rate ($P_{rev}$) = 1, Total power coefficient ($\zeta$) = 0.1 |
| PSO[*][14] | Inertia weight damping ratio = 0.99, Personal learning coefficient (c1) = 1.5, Global learning coefficient (c2) = 2, $Vel_{max}$ = +0.1($Var_{max}$ -$Var_{min}$), $Vel_{min}$ = - $Vel_{max}$ using Eq. (30) |
| MSA[*][37] | Number of new solutions = 20 (#new Neighbors), Mutation rate (m) = 0.5, Maximum number of sub-iterations = 20, Initial temperature adjustment based on Medina's method [89],[90] = 1/20 * Cost of initial best solution, Cooling coefficient (k) = 0.85 |
| IHS[*][35] | Number of new solutions = 20 (#new Harmonies), Harmony Memory Size (HMS) = 50, Harmony Memory = 0.85, Dynamic Pitch Adjustment Rate (PAR) [35],[36] $PAR_{max}$ = 0.99, $PAR_{min}$ = 0.35, Dynamic Bandwidth (BW)[36] $BW_{max}$ = 4, $BW_{min}$ = 1e-5, Number of new Harmonies = 20 |
| FAGLSUD | Number of sub-populations = 20 (#empires), $Vel_{max}$ and $Vel_{min}$ using Eq. (30) and (37) |

* We included the results of the algorithms marked with the star sign for the same optimization problem landscape, assuming twenty-six constraints and twelve design variables. In addition to the newly studied methods, some standard metaheuristics, such as ICA, DE and PSO, were used for a comprehensive comparison because of their similarity to our method. For BBO and DE, we directly reported the low-cost and low-weight design results under nine pseudo-static loading cases achieved in the study conducted by Gandomi et al. [49].

## 7.1 Design examples

The utilized design parameters for examples 1 and 2 are indicated in Table 9. Each design example was analyzed under nine combinations of horizontal and vertical seismic coefficients between 0 and 0.3, as listed in Table 10. The means of the final low-cost, low-weight, and low-

$CO_2$ design objective function values of examples 1 and 2 are reported in Table 11 and Table 12, respectively, which include the SD values of each objective. After proving the superiority of the FAGLSUD over other RCC optimization metaheuristics using the mean objective values, Appendix A and B provide the lowest values of the final low-cost, low-weight, and low-$CO_2$ retaining wall design variables along with their associated objective values for examples 1 and 2 under 9 different seismic conditions; see Table A 1 , Table A 2, Table A 3, Table B 1, Table B 2, and Table B 3. All design solutions and objective results are rounded to the nearest two decimal places.

Table 9: The design parameters for two retaining wall examples

| Design Parameter | Unit | Symbol | Example 1 | Example 2 |
|---|---|---|---|---|
| Height of stem | m | $H$ | 3 | 6 |
| Yield strength of reinforcing steel | MPa | $f_y$ | 400 | 400 |
| Compressive strength of concrete | MPa | $f_c$ | 21 | 21 |
| Concrete cover | m | $C_c$ | 0.07 | 0.07 |
| Shrinkage and temperature reinforcement percent | - | $\rho_{st}$ | 0.002 | 0.002 |
| Surcharge load | kPa | $q$ | 20 | 30 |
| Backfill slope | degree | $i$ | 10 | 0 |
| Internal friction angle of retained soil | degree | $\phi$ | 36 | 32 |
| Internal friction angle of base soil | degree | $\phi'$ | 36 | 32 |
| Unit weight of retained soil | kN/m³ | $\gamma_s$ | 17.5 | 20 |
| Unit weight of base soil | kN/m³ | $\gamma'_s$ | 18.5 | 18 |
| Unit weight of concrete | kN/m³ | $\gamma_c$ | 23.5 | 23.5 |
| Cohesion of base soil | kPa | $c$ | 125 | 100 |
| Depth of soil in front of wall | m | $D$ | 0.5 | 1 |
| Cost of steel | $/kg | $C_s$ | 0.4 | 0.4 |
| Cost of concrete | $/m³ | $C_c$ | 40 | 40 |
| Desirable Factor of safety for overturning stability | - | $FS_{O,desired}$ | 1.5 | 1.5 |
| Desirable Factor of safety against sliding | - | $FS_{S,desired}$ | 1.5 | 1.5 |
| Desirable Factor of safety for bearing capacity | - | $FS_{B,desired}$ | 3 | 3 |
| Emissions of steel | Kg/kg | $e_s$ | 2.82 | 2.82 |
| Emissions of concrete | Kg/m³ | $e_c$ | 224.94 | 224.94 |

Table 10: Nine combinations of seismic coefficients

| Case study | Seismic coefficient | |
|---|---|---|
| | $k_h$ | $k_v$ |
| 1 | 0 | 0 |
| 2 | 0.15 | 0 |
| 3 | 0.3 | 0 |
| 4 | 0 | 0.15 |
| 5 | 0.15 | 0.15 |
| 6 | 0.3 | 0.15 |
| 7 | 0 | 0.3 |
| 8 | 0.15 | 0.3 |
| 9 | 0.3 | 0.3 |

## 7.2 Comparison with prospering algorithms in the literature community

In this section, FAGLSUD is compared with the most recently successful EAs (LFBBO, BBO), SIs (FPA, GWO), and PSs (MSA, IHS) reported in literature in terms of computational performance. The conventional EAs (DE) and SIs (ICA and PSO) were included in our comparisons due to similarities with our proposed method. The results confirm that the establishment of innovative search mechanisms along with fuzzy-based parameter adaptation and operator selection in FAGLSUD is effective to locate global optimal solution in a large-sized design space.

### 7.2.1 Example 1: Design of 3m-tall retaining wall with minimum cost, weight, and $CO_2$ emission

This example was designed thrice using each method, once with a low-cost, once with a low-weight, and once with a low-$CO_2$ objective function value. Seismic performance of each design was checked against earthquake disasters using the nine different pseudo-static coefficients denoted as loading Cases 1 to 9. By considering the horizontal coefficient ($k_h$) of 0 (Case 1, 4, and 7), it can be concluded that increasing vertical coefficient ($k_v$) from 0 to 0.3 causes decrements in the final cost, weight, and $CO_2$ results. A similar pattern on decreasing the final design results is demonstrated by considering a $k_h$ of 0.15 (Case 2, 5, and 8) and increasing $k_v$ from 0 to 0.3. However, this trend is reversed by switching the horizontal loading condition (Cases 1-3, and 4-6), so that increasing $k_h$ from 0 to 0.3 causes increments in final cost, weight, and $CO_2$ results. Finally, the most severe loading condition was experienced with both $k_h$ and $k_v$ equal to 0.3, for which our method performed better than the other algorithms.

According to the first loading case (static: $k_v$ and $k_h$ = 0) results in Table 11, after 101 independent runs, the average cost obtained by FAGLSUD was \$62.45/m and the coefficient of variation was 0.11. FAGLSUD, MSA, PSO, and LFBBO are the four competitive algorithms that obtained the best, second, third, and fourth-best average cost, weight and $CO_2$ in most of the cases. In comparison, the FAGLSUD design has around 0.08%, 0.65% and 1.14% less average cost than MSA, PSO and LFBBO, respectively, 0.004%, 0.21%, and 0.004% lighter average weight than MSA, PSO and LFBBO, respectively; and 0.005%, 0.24% and 1.68% lower average $CO_2$ emission than MSA, PSO and LFBBO, respectively. This trend continues for most of the simulation results reported in Table 11 and Table 12, where FAGLSUD showed the best mean cost, weight, and $CO_2$ values in most cases. The mean solutions were obtained based on averages of the best solutions over 101 independent runs. The dark grey highlighted values imply the best performance, and light grey highlighted values imply the second-best performance. It is worth noting that, while these enhancements may look insignificant per wall meter, they become important for walls that extend along several kilometers in road construction. In addition, the dimensions of the optimization problem at hand, the number of iterations and the initial solutions are important factors that affect the final enhancements (accuracy improvements). As discussed earlier, we repeated each of the algorithms 1000 times to ensure enough accuracy for the comparison methods. Therefore, instead of comparing the methods using the best of 101 independent runs, we compared the methods using the mean of 101 separate runs. We used random initial solutions, which might make the initial stages of the optimization process inefficient owing to the infeasibility of candidate solutions in the population; a potential solution is to design the optimization process to start from a set of feasible solutions (see [79]).

We used the Friedman ranks for all algorithms. As is evident from ranking values at the bottom of each mean objective value in Table 11 and Table 12, in all six sets of case observations (each set involves nine case observations; case 1 to 9), the FAGLSUD clearly provides, and by a wide margin, the lowest average rank and therefore the best overall rank. The superiority of FAGLSUD on average results over a series of 101 runs confirms the superior performance of our proposed approach. Among other algorithms, the MSA was second best in 5 of the 6 sets of the case observations, the PSO was third best in 4 sets of the case observations and the LFBBO ranked fourth-best overall in 1 set of observations. In the next section, we also employ a statistical Wilcoxon signed-rank test for a pairwise comparison between FAGLSUD and the other metaheuristics, and report the symbol † over each set of nine case observations.

The computational performance of each algorithm was evaluated according to convergence rates, as illustrated in Appendix A (Figure A 1, Figure A 2, Figure A 3), and Appendix B (Figure B 1, Figure B 2, and Figure B 3), for examples 1 and 2, respectively. The convergence rate plots are based on mean iteration results over 101 series runs. Some methods such as the MSA have high local search (exploitation), which clearly enhances their convergence rate, but without improving the accuracy at the end of convergence time. Consequently, fast convergence tendency can significantly reduce the global search capability of the algorithm and increases the likelihood of the search ending up at a local minimum. On the other hand, as evident from Table 11 and Table 12, such fast methods (e.g., MSA) do not offer superior accuracy and stability, which indicates two issues: they fall into local optima and are sensitive to random initial solutions. With regards to stability of the results, LFBBO, almost in all observations, shows the lowest sensitivity to initialization (i.e., lowest SD), but it suffers from both low accuracy and low convergence rate. In contrast, it is clear that FAGLSUD approaches the minimum average cost, weight and $CO_2$ almost exponentially with the number of iterations, it is more stable and quicker than most of the algorithms. The powerful global search (exploration) avoids trapping in local optimal solutions, which clearly increases the accuracy, but also decreases the convergence rate during early and intermediate stages of convergence. The balance between global search, local search and information sharing, using appropriate setting of parameters and a suitable selection of the most prominent operators at each optimization stage, contributes to the consistent behavior in terms of accuracy, stability, and convergence time across various case observations. The superior performance in terms of accuracy will be further confirmed by Friedman test ranks followed by Wilcoxon statistical analysis in the next section.

Table 11: Cost, Weight and $CO_2$ optimization results for Example 1

| Case | Obj. | Met. | SIs | | | | EAs | | | PSs | | SI and EA |
|---|---|---|---|---|---|---|---|---|---|---|---|---|
| | | | FPA† *[2] | GWO† *[41] | ICA† *[17] | PSO† *[14] | LFBBO† *[38] | BBO† *[49] | DE† *[47] | MSA† *[37] | IHS† *[35] | Proposed FAGLSUD |
| 1 | Cost (USD) | Mean | 83.66 | 86.10 | 73.84 | 62.86 | 63.16 | 62.96 | 86.06 | 62.50 | 63.80 | **62.45** |
| | | Rank | 8 | 10 | 7 | 3 | 5 | 4 | 9 | 2 | 6 | **1** |
| | | SD | 4.90 | 3.73 | 4.74 | 0.50 | **0** | 0.35 | 3.58 | 0.16 | 0.76 | 0.11 |
| 2 | Cost (USD) | Mean | 97.78 | 102.94 | 91.06 | 84.01 | 89.08 | 84.65 | 102.86 | 83.98 | 84.34 | **83.93** |
| | | Rank | 8 | 9 | 7 | 3 | 6 | 5 | 10 | 2 | 4 | **1** |
| | | SD | 2.48 | 3.22 | 2.54 | 0.38 | **0** | 1.32 | 3.41 | 0.37 | 0.74 | 0.39 |
| 3 | Cost (USD) | Mean | 133.92 | 140.60 | 126.17 | 117.36 | 121.31 | 118.84 | 140.97 | 116.97 | 118.39 | **116.71** |
| | | Rank | 8 | 9 | 7 | 3 | 6 | 5 | 10 | 2 | 4 | **1** |

|  |  |  | FPA† *[2] | GWO† *[41] | ICA† *[17] | PSO† *[14] | LFBBO† *[38] | BBO† *[49] | DE† *[47] | MSA† *[37] | IHS† *[35] | FAGLSUD |
|---|---|---|---|---|---|---|---|---|---|---|---|---|
|  |  | SD | 3.37 | 3.77 | 3.46 | 0.79 | **0** | 1.81 | 3.90 | 0.56 | 1.31 | 0.49 |
| 4 | Cost (USD) | Mean | 81.71 | 84.15 | 72.64 | 59.78 | 59.37 | 59.36 | 84.94 | 59.38 | 60.52 | **59.35** |
|  |  | Rank | 8 | 9 | 7 | 5 | 3 | 2 | 10 | 4 | 6 | **1** |
|  |  | SD | 5.23 | 4.17 | 5.18 | 0.97 | 0.78 | **0.10** | 3.21 | 0.80 | 0.85 | 0.77 |
| 5 | Cost (USD) | Mean | 92.89 | 97.79 | 86.13 | **79.51** | 82.33 | 80.19 | 98.57 | 79.62 | 79.87 | **79.51** |
|  |  | Rank | 8 (7) | 9 (8) | 7 (6) | **1.5 (1)** | 6 (5) | 5 (4) | 10 (9) | 3 (2) | 4 (3) | **1.5 (1)** |
|  |  | SD | 2.55 | 3.13 | 2.91 | 0.46 | **0** | 1.03 | 3.18 | 0.46 | 0.60 | 0.41 |
| 6 | Cost (USD) | Mean | 136.73 | 143.03 | 128.12 | 119.63 | 121.33 | 121.26 | 142.84 | 119.00 | 120.75 | **118.90** |
|  |  | Rank | 8 | 10 | 7 | 3 | 6 | 5 | 9 | 2 | 4 | **1** |
|  |  | SD | 4.01 | 3.43 | 3.23 | 0.75 | **0** | 1.72 | 3.94 | 0.46 | 1.22 | 0.45 |
| 7 | Cost (USD) | Mean | 81.62 | 84.08 | 71.09 | 57.15 | 56.89 | 56.96 | 84.19 | 56.89 | 57.83 | **56.88** |
|  |  | Rank | 8 (7) | 9 (8) | 7 (6) | 5 (4) | 2.5 (2) | 4 (3) | 10 (9) | 2.5 (2) | 6 (5) | **1** |
|  |  | SD | 5.32 | 3.88 | 5.37 | 0.44 | **0.01** | 0.07 | 3.66 | **0.01** | 0.76 | **0.01** |
| 8 | Cost (USD) | Mean | 89.53 | 94.16 | 82.00 | 75.49 | 77.21 | 75.76 | 94.11 | 75.67 | 75.60 | **75.44** |
|  |  | Rank | 8 | 10 | 7 | 2 | 6 | 5 | 9 | 4 | 3 | **1** |
|  |  | SD | 3.24 | 3.30 | 2.54 | 0.26 | **0** | 0.62 | 3.21 | 0.19 | 0.47 | 0.12 |
| 9 | Cost (USD) | Mean | 152.26 | 157.01 | 142.61 | 131.80 | 131.83 | 133.90 | 157.75 | 131.89 | 133.05 | **131.06** |
|  |  | Rank | 8 | 9 | 7 | 2 | 3 | 6 | 10 | 4 | 5 | **1** |
|  |  | SD | 4.88 | 4.77 | 3.81 | 0.73 | **0** | 1.57 | 4.19 | 0.79 | 1.27 | 0.35 |
| Average rank |  |  | 8 | 9.33 | 7 | 3.06 | 4.83 | 4.56 | 9.67 | 2.83 | 4.67 | **1.05** |
| Overall rank |  |  | 8 | 9 | 7 | 3 | 6 | 4 | 10 | 2 | 5 | **1** |
|  |  |  | FPA† *[2] | GWO† *[41] | ICA† *[17] | PSO† *[14] | LFBBO† *[38] | BBO† *[49] | DE† *[47] | MSA† *[37] | IHS† *[35] | FAGLSUD |
| 1 | Weight (Kg) | Mean | 2911.51 | 3033.61 | 2711.68 | 2487.09 | 2481.96 | 2485.27 | 3023.22 | 2481.97 | 2513.68 | **2481.87** |
|  |  | Rank | 8 | 10 | 7 | 5 | 2 | 4 | 9 | 3 | 6 | **1** |
|  |  | SD | 97.25 | 98.47 | 79.30 | 7.80 | **2.73** | 3.60 | 82.70 | 2.82 | 25.79 | 2.78 |
| 2 | Weight (Kg) | Mean | 3356.98 | 3504.69 | 3162.98 | 3058.40 | 3124.30 | 3069.97 | 3481.43 | 3055.42 | 3068.44 | **3055.33** |
|  |  | Rank | 7 | 9 | 6 | 3 | 6 | 5 | 8 | 2 | 4 | **1** |
|  |  | SD | 62.09 | 83.41 | 50.63 | 5.14 | **0** | 24.16 | 89.02 | 1.98 | 12.90 | 1.90 |
| 3 | Weight (Kg) | Mean | 4455.24 | 4679.61 | 4281.72 | 4211.16 | 4367.50 | 4264.64 | 4697.38 | 4209 | 4219.24 | **4208.16** |
|  |  | Rank | 8 | 9 | 6 | 3 | 7 | 5 | 10 | 2 | 4 | **1** |
|  |  | SD | 57.42 | 96.67 | 37.21 | 6.53 | **0** | 49.01 | 95.07 | 3.09 | 12.75 | 2.74 |
| 4 |  | Mean | 2920.89 | 3026.56 | 2705.96 | 2477.83 | 2472.63 | 2474.98 | 3027.84 | 2472.53 | 2509.22 | **2472.46** |

| | | | FPA†*[2] | GWO†*[41] | ICA†*[17] | PSO†*[14] | LFBBO†*[38] | BBO†*[49] | DE†*[47] | MSA†*[37] | IHS†*[35] | FAGLSUD |
|---|---|---|---|---|---|---|---|---|---|---|---|---|
| | Weight (Kg) | Rank | 8 | 9 | 7 | 5 | 3 | 4 | 10 | 2 | 6 | **1** |
| | | SD | 120.06 | 82.72 | 100.67 | 11.22 | 4.37 | 4.03 | 85.77 | 3.81 | 29.59 | **3.20** |
| 5 | Weight (Kg) | Mean | 3235.50 | 3360.49 | 3042.12 | 2953.54 | 2952.64 | 2958.39 | 3379.01 | 2951.20 | 2955.84 | **2950.91** |
| | | Rank | 8 | 9 | 7 | 4 | 3 | 6 | 10 | 2 | 5 | **1** |
| | | SD | 77.98 | 91.32 | 49.42 | 5.26 | **0** | 23.59 | 83.83 | 0.61 | 5.35 | 0.52 |
| 6 | Weight (Kg) | Mean | 4560.24 | 4795.98 | 4374.04 | 4312.14 | 4520.78 | 4361.07 | 4799.30 | 4586.98 | 4586.96 | **4308.41** |
| | | Rank | 6 | 8 | 4 | 2 | 5 | 3 | 9 | 7 | 6 | **1** |
| | | SD | 31.73 | 115.08 | 28.02 | 8.56 | **0** | 46.96 | 103.87 | 24.23 | 25.40 | 3.29 |
| 7 | Weight (Kg) | Mean | 2895.98 | 2999.95 | 2707.38 | 3058.40 | 2465.45 | 2469.15 | 2999.92 | 2466.45 | 2498.59 | **2465.38** |
| | | Rank | 7 | 9 | 6 | 10 | 2 | 4 | 8 | 3 | 5 | **1** |
| | | SD | 122.77 | 93.52 | 95.98 | 5.14 | 4.41 | 3.33 | 88.92 | 4.41 | 29.23 | **2.14** |
| 8 | Weight (Kg) | Mean | 3155.78 | 3252.52 | 2963.35 | 3058.40 | 2858.56 | 2860.20 | 3263.75 | 2858.70 | 2859.92 | **2858.19** |
| | | Rank | 8 | 9 | 6 | 7 | 2 | 5 | 10 | 3 | 4 | **1** |
| | | SD | 76.69 | 79.14 | 61.14 | 5.14 | 0.57 | 2.60 | 80.32 | 1.44 | 2.61 | **0.25** |
| 9 | Weight (Kg) | Mean | 5063.88 | 5270.24 | 4876.80 | 4789.49 | 4876.12 | 4815.37 | 5267.65 | 4782.02 | 4796.98 | **4781.25** |
| | | Rank | 8 | 10 | 7 | 3 | 6 | 5 | 9 | 2 | 4 | **1** |
| | | SD | 80.53 | 102.94 | 61.77 | 19.44 | **0** | 40.35 | 81.36 | 5.05 | 18.82 | 4.16 |
| Average rank | | | 7.56 | 9.11 | 6.22 | 4.67 | 4 | 4.56 | 9.22 | 2.89 | 4.89 | **1** |
| Overall rank | | | 8 | 9 | 7 | 5 | 3 | 4 | 10 | 2 | 6 | **1** |
| | | | FPA†*[2] | GWO†*[41] | ICA†*[17] | PSO†*[14] | LFBBO†*[38] | BBO†*[49] | DE†*[47] | MSA†*[37] | IHS†*[35] | FAGLSUD |
| 1 | $CO_2$ (Kg) | Mean | 506.33 | 533.48 | 454.07 | 380.88 | 386.36 | **379.97** | 383.84 | 379.99 | 389.66 | **379.97** |
| | | Rank | 9 (8) | 10 (9) | 8 (7) | 4 (3) | 6 (5) | **1.5 (1)** | 5 (4) | 3 (2) | 7 (6) | **1.5 (1)** |
| | | SD | 30.74 | 21.44 | 28.67 | 2.37 | **0** | 1.63 | 4.11 | 1.64 | 5.14 | 1.62 |
| 2 | $CO_2$ (Kg) | Mean | 602.95 | 634.73 | 557.23 | 513.31 | 556.09 | 517.21 | 512.67 | 512.80 | 516.31 | **512.66** |
| | | Rank | 9 | 10 | 8 | 4 | 7 | 6 | 2 | 3 | 5 | **1** |
| | | SD | 18.01 | 22.44 | 14.75 | 2.71 | **0** | 6.28 | 1.93 | 3.31 | 4.85 | 1.74 |
| 3 | $CO_2$ (Kg) | Mean | 833.45 | 870.85 | 783.59 | 721.6 | 800.85 | 733.26 | 723.16 | 712.79 | 728.00 | **712.77** |
| | | Rank | 9 | 10 | 7 | 3 | 8 | 6 | 4 | 2 | 5 | **1** |
| | | SD | 20.89 | 24.71 | 21.80 | 7.42 | **0** | 9.95 | 5.98 | 3.27 | 9.01 | 3.27 |
| 4 | $CO_2$ (Kg) | Mean | 505.91 | 521.27 | 436.19 | 361.17 | 359.96 | 359.56 | 361.74 | 359.80 | 365.73 | **359.07** |

|   |   |       | 1 | 2 | 3 | 4 | 5 | 6 | 7 | 8 | 9 | 10 |
|---|---|-------|---|---|---|---|---|---|---|---|---|----|
|   |   | Rank  | 9 | 10 | 8 | 5 | 4 | 2 | 6 | 3 | 7 | **1** |
|   |   | SD    | 28.59 | 22.34 | 29.76 | 2.97 | **1.13** | 1.60 | 3.78 | 1.42 | 5.08 | 1.34 |
| 5 | $CO_2$ (Kg) | Mean | 575.21 | 609.52 | 529.00 | 485.55 | 510.76 | 487.58 | 484.03 | 484.99 | 488.78 | **484.01** |
|   |   | Rank  | 9 | 10 | 8 | 4 | 7 | 5 | 2 | 3 | 6 | **1** |
|   |   | SD    | 20.72 | 21.47 | 19.08 | 3.52 | **0** | 4.82 | 1.72 | 3.84 | 5.34 | 3.22 |
| 6 | $CO_2$ (Kg) | Mean | 851.48 | 889.46 | 797.41 | 734.68 | 741.45 | 745.96 | 736.40 | 728.84 | 744.22 | **728.47** |
|   |   | Rank  | 9 | 10 | 8 | 3 | 5 | 7 | 4 | 2 | 6 | **1** |
|   |   | SD    | 23.71 | 23.48 | 20.60 | 7.73 | **0** | 11.22 | 5.52 | 2.74 | 9.31 | 2.23 |
| 7 | $CO_2$ (Kg) | Mean | 504.23 | 513.59 | 439.92 | 344.07 | 342.18 | 343.44 | 344.21 | 342.19 | 348.32 | **342.11** |
|   |   | Rank  | 9 | 10 | 8 | 5 | 2 | 4 | 6 | 3 | 7 | **1** |
|   |   | SD    | 32.32 | 24.98 | 34.38 | 1.60 | **0.19** | 2.42 | 3.66 | 0.21 | 4.34 | **0.19** |
| 8 | $CO_2$ (Kg) | Mean | 555.91 | 585.50 | 505.00 | 459.48 | 473.40 | 460.37 | 459.80 | 459.38 | 462.08 | **459.37** |
|   |   | Rank  | 9 | 10 | 8 | 3 | 7 | 5 | 4 | 2 | 6 | **1** |
|   |   | SD    | 22.11 | 17.65 | 17.16 | 2.34 | **0** | 2.73 | 1.20 | 3.5 | 4.25 | 2.25 |
| 9 | $CO_2$ (Kg) | Mean | 963.07 | 979.52 | 890.85 | 810.16 | 808.68 | 825.72 | 816.39 | 805.64 | 822.12 | **805.37** |
|   |   | Rank  | 9 | 10 | 8 | 4 | 3 | 7 | 5 | 2 | 6 | **1** |
|   |   | SD    | 41.09 | 28.51 | 25.84 | 5.43 | **0** | 10.24 | 8.17 | 1.94 | 10.84 | 1.76 |
| **Average rank** | | | 9 | 10 | 7.89 | 4.11 | 5.44 | 4.83 | 4.22 | 2.56 | 6 | **1.06** |
| **Overall rank** | | | 9 | 10 | 8 | 3 | 6 | 5 | 4 | 2 | 7 | **1** |

### 7.2.2 Example 2: Design of 6m-tall retaining wall with minimum cost, weight, and $CO_2$ emission

The second design example represents a more intensified level of complexity by enhancing the height of the wall. The second design is utilized not just to explore the efficiency of the proposed algorithm for low-cost, weight, and $CO_2$ design, but also to check the effect of the height of the wall and surcharge load on the final design under nine seismic loading analyses. Comparing example 2 with 1, it can be concluded that the differences in costs, weight, and $CO_2$ achieved by 10 metaheuristics become more significant for large wall heights, where high costs, weights, and $CO_2$ emission are generally involved.

In this example, the low-cost and low-$CO_2$ design will experience only minor effects from an earthquake with less difference between each seismic loading case and the following loading case. However, the low-weight design showed more sensitivity to the variation of earthquake coefficients. Similar to example 1, there are two trends in all the cases: 1) increasing the vertical earthquake coefficient leads to a more economical, lighter, and more sustainable design, and 2) enhancing the horizontal coefficient causes the final designs to become less economical, sustainable, and heavier.

Based on Table 12, in most of the cases, FAGLSUD reached the lowest mean optimization results for low-cost, low-weight, and low-$CO_2$ designs. In case 1, the FAGLSUD design had 0.16%, 0.19% and 0.38% less average cost, 0.08%, 0.04% and 5.21% lighter average weight, and 0.04%, 0.17% and 2.09% less $CO_2$ (greater sustainability) than MSA, PSO and LFBBO, respectively. As discussed for example 1, these small enhancements will be significant for walls that extend along several kilometers in road construction, or for higher levels of complexity (e.g., with an increase in the number of dimensions), or when feasible solutions are utilized in the initial population [79].

The mean optimization stories over 101 independent runs are depicted in Figure B 1, Figure B 2, and Figure B 3 in Appendix B. FAGLSUD, with a reasonable convergence speed, converges to the minimum average cost, weight, and $CO_2$ towards the end of the convergence time, while the faster methods easily get stuck in local optima. This result is owed to the relatively superior exploration capability of FAGLSUD, whose iteration histories follow an exponential trend in all pseudo-static conditions and show more predictable results than the other approaches. Finally, FAGLSUD converges to more accurate results than the other algorithms, which can be attributed to its relatively excellent exploration in the early stage of iterations as well as its excellent exploitation and information sharing capacity in the later stage of the iterations. Considering average cost, weight, and $CO_2$, coefficients of variation, and convergence trend and rates, FAGLSUD provides better accuracy with superior speed and stability compared to many of the other methods. Table B 1, Table B 2, and Table B 3 in Appendix B provide the low-cost, low-weight, and low-$CO_2$ design solutions, respectively.

Table 12: Cost, weight and $CO_2$ optimization results for example 2

| Case | Obj. | Met. | SIs | | | | EAs | | | PSs | | SI and EA |
|---|---|---|---|---|---|---|---|---|---|---|---|---|
| | | | FPA† *[2] | GWO† *[41] | ICA† *[17] | PSO† *[14] | LFBBO† *[38] | BBO† *[49] | DE† *[47] | MSA† *[37] | IHS† *[35] | Proposed FAGLSUD |
| 1 | Cost (USD) | Mean | 276.20 | 299.13 | 281.11 | 247.85 | 248.33 | 381.89 | 432.52 | 247.78 | 250.21 | **247.39** |
| | | Rank | 6 | 8 | 7 | 3 | 4 | 9 | 10 | 2 | 5 | **1** |
| | | SD | 6.00 | 8.38 | 8.34 | 1.01 | **0** | 2.84 | 10.64 | 0.02 | 1.66 | 1.03 |
| 2 | Cost (USD) | Mean | 329.09 | 355.27 | 326.32 | 311.74 | 315.27 | 382.33 | 429.79 | 312.75 | 313.56 | **311.00** |
| | | Rank | 7 | 8 | 6 | 2 | 5 | 9 | 10 | 3 | 4 | **1** |
| | | SD | 3.75 | 8.55 | 6.05 | 0.85 | **0** | 5.30 | 10.85 | 1.36 | 2.22 | 1.27 |
| 3 | Cost (USD) | Mean | 429.54 | 444.49 | 400.23 | 384.26 | 385.75 | 382.44 | 431.05 | 383.98 | 386.51 | **382.43** |
| | | Rank | 8 | 10 | 7 | 4 | 5 | 2 | 9 | 3 | 6 | **1** |
| | | SD | 17.50 | 11.83 | 9.22 | 1.06 | **0** | 3.57 | 9.73 | **0** | 3.14 | 0.77 |
| 4 | Cost (USD) | Mean | 258.61 | 281.66 | 262.24 | 247.85 | 229.42 | 383.33 | 430.34 | 229.80 | 232.27 | **229.41** |
| | | Rank | 6 | 8 | 7 | 5 | 2 | 9 | 10 | 3 | 4 | **1** |
| | | SD | 5.77 | 8.56 | 8.68 | 1.06 | **0** | 7.33 | 11.24 | 1.23 | 1.70 | 1.04 |
| 5 | Cost (USD) | Mean | 309.53 | 335.33 | 310.64 | 292.38 | 294.03 | 383.00 | 430.70 | 293.28 | 293.96 | **292.34** |
| | | Rank | 6 | 8 | 7 | 2 | 5 | 9 | 10 | 3 | 4 | **1** |
| | | SD | 4.04 | 7.34 | 8.04 | 1.12 | **0** | 6.87 | 11.95 | 1.58 | 2.28 | 1.09 |

| | | | FPA† *[2] | GWO† *[41] | ICA† *[17] | PSO† *[14] | LFBBO† *[38] | BBO† *[49] | DE† *[47] | MSA† *[37] | IHS† *[35] | FAGLSUD |
|---|---|---|---|---|---|---|---|---|---|---|---|---|
| 6 | Cost (USD) | Mean | 410.29 | 432.56 | 392.00 | 377.50 | 378.33 | 382.86 | 429.84 | 376.57 | 380.17 | **376.53** |
| | | Rank | 8 | 10 | 7 | 3 | 4 | 6 | 9 | 2 | 5 | **1** |
| | | SD | 11.80 | 11.86 | 8.71 | 1.95 | **0** | 6.56 | 9.31 | 2.46 | 2.01 | 1.54 |
| 7 | Cost (USD) | Mean | 239.84 | 262.14 | 241.10 | 207.69 | 207.84 | 382.27 | 432.38 | 207.93 | 209.99 | **207.56** |
| | | Rank | 6 | 8 | 7 | 2 | 3 | 9 | 10 | 4 | 5 | **1** |
| | | SD | 6.44 | 8.59 | 9.41 | 0.66 | **0** | 5.76 | 10.58 | 0.50 | 1.62 | 0.39 |
| 8 | Cost (USD) | Mean | 288.95 | 315.68 | 292.31 | 271.73 | 274.32 | 382.84 | 431.90 | 271.64 | 272.88 | **271.40** |
| | | Rank | 6 | 8 | 7 | 3 | 5 | 9 | 10 | 2 | 4 | **1** |
| | | SD | 3.94 | 7.25 | 8.97 | 0.93 | **0** | 5.00 | 10.54 | 1.22 | 1.85 | 0.92 |
| 9 | Cost (USD) | Mean | 416.75 | 430.95 | 393.75 | 378.88 | 380.09 | 382.33 | 432.74 | 378.77 | 381.73 | **378.02** |
| | | Rank | 8 | 9 | 7 | 3 | 4 | 6 | 10 | 2 | 5 | **1** |
| | | SD | 15.43 | 12.84 | 9.82 | 0.91 | **0** | 4.94 | 11.37 | 0.62 | 2.21 | 0.48 |
| **Average rank** | | | 6.78 | 8.56 | 6.89 | 3 | 4.11 | 7.56 | 9.78 | 2.67 | 4.67 | **1** |
| **Overall rank** | | | 6 | 9 | 7 | 3 | 4 | 8 | 10 | 2 | 5 | **1** |
| | | | FPA† *[2] | GWO† *[41] | ICA† *[17] | PSO† *[14] | LFBBO† *[38] | BBO† *[49] | DE† *[47] | MSA† *[37] | IHS† *[35] | FAGLSUD |
| 1 | Weight (Kg) | Mean | 9345.19 | 10047.63 | 9485.10 | 8644.37 | 9091.10 | 8847.05 | 9980.74 | 8647.37 | 8734.45 | **8640.82** |
| | | Rank | 7 | 10 | 8 | 2 | 6 | 5 | 9 | 3 | 4 | **1** |
| | | SD | 200.05 | 254.06 | 347.38 | 10.72 | **0** | 181.33 | 216.58 | **0** | 57.23 | 5.85 |
| 2 | Weight (Kg) | Mean | 10946.14 | 11640.84 | 10841.56 | 10709.76 | 10744.01 | 10857.59 | 11558.74 | 10718.52 | 10729.16 | **10709.61** |
| | | Rank | 8 | 10 | 6 | 2 | 5 | 7 | 9 | 3 | 4 | **1** |
| | | SD | 112.49 | 213.62 | 120.69 | 4.43 | **0** | 170.24 | 169.64 | 0.82 | 23.47 | 4.12 |
| 3 | Weight (Kg) | Mean | 14696.29 | 14302.05 | 13710.74 | 13325.32 | 13439.86 | 13554.98 | 14221.28 | 13323.96 | 13349.10 | **13323.81** |
| | | Rank | 10 | 9 | 7 | 3 | 5 | 6 | 8 | 2 | 4 | **1** |
| | | SD | 427.91 | 246.26 | 347.75 | 6.49 | **0** | 434.99 | 214.79 | 0.45 | 65.41 | 0.99 |
| 4 | Weight (Kg) | Mean | 9028.42 | 9630.05 | 9203.00 | 8281.45 | 9024.82 | 8544.09 | 9600.95 | 8277.76 | 8367.31 | **8277.24** |
| | | Rank | 7 | 10 | 8 | 3 | 6 | 5 | 9 | 2 | 4 | **1** |
| | | SD | 229.46 | 250.14 | 343.73 | 8.53 | **0** | 235.87 | 230.61 | 4.37 | 56.39 | 4.27 |
| 5 | Weight (Kg) | Mean | 10349.34 | 11038.28 | 10319.87 | **10086.42** | 10186.39 | 10253.45 | 10974.28 | 10089.27 | 10105.21 | 10087.33 |
| | | Rank | 8 | 10 | 7 | 1 | 5 | 6 | 9 | 3 | 4 | 2 |

|   |   |   | FPA[2] | GWO[41] | ICA[17] | PSO[14] | LFBBO[38] | BBO[49] | DE[47] | MSA[37] | IHS[35] | FAGLSUD |
|---|---|---|---|---|---|---|---|---|---|---|---|---|
|   |   | SD | 121.27 | 210.67 | 194.45 | 8.87 | **0** | 177.44 | 196.36 | 8.94 | 25.32 | 7.89 |
| 6 | Weight (Kg) | Mean | 13772.12 | 14000.83 | 13354.71 | 13105.54 | 13196.88 | 13191.09 | 13971.36 | 13105.60 | 13113.86 | **13104.87** |
|   |   | Rank | 8 | 10 | 7 | 2 | 6 | 5 | 9 | 3 | 4 | **1** |
|   |   | SD | 346.54 | 197.47 | 252.75 | 4.88 | **0** | 168.42 | 178.33 | 4.68 | 14.05 | 3.85 |
| 7 | Weight (Kg) | Mean | 8842.71 | 9371.35 | 8883.26 | 8098.19 | 8093.70 | 8129.49 | 9384.10 | 8093.70 | 8166.43 | **8093.61** |
|   |   | Rank | 7 (6) | 9 (8) | 8 (7) | 4 (3) | 2.5 (2) | 5 (4) | 10 (8) | 2.5 (2) | 6 (5) | **1** |
|   |   | SD | 205.75 | 229.57 | 339.39 | 7.02 | 2.56 | 9.61 | 210.35 | 3.01 | 62.06 | **2.55** |
| 8 | Weight (Kg) | Mean | 9784.39 | 10508.09 | 9817.37 | **9444.05** | 10486.56 | 9609.26 | 10435.38 | 9449.88 | 9466.99 | 9444.77 |
|   |   | Rank | 6 | 10 | 7 | **1** | 9 | 5 | 8 | 3 | 4 | 2 |
|   |   | SD | 157.04 | 248.44 | 250.91 | 6.00 | **0** | 180.59 | 195.48 | 6.81 | 31.84 | 6.57 |
| 9 | Weight (Kg) | Mean | 13865.58 | 14113.29 | 13377.52 | 13180.74 | 13260.20 | 13296.46 | 14044.45 | 13179.35 | 13190.45 | **13179.25** |
|   |   | Rank | 8 | 10 | 7 | 3 | 5 | 6 | 9 | 2 | 4 | **1** |
|   |   | SD | 397.84 | 213.19 | 207.32 | 7.29 | **0** | 280.70 | 181.95 | 4.52 | 16.51 | 4.48 |
| Average rank | | | 7.67 | 9.78 | 7.22 | 2.33 | 5.5 | 5.44 | 8.89 | 2.61 | 4.22 | **1.22** |
| Overall rank | | | 8 | 10 | 7 | 2 | 6 | 5 | 9 | 3 | 4 | **1** |
|   |   |   | FPA†*[2] | GWO†*[41] | ICA†*[17] | PSO†*[14] | LFBBO†*[38] | BBO†*[49] | DE†*[47] | MSA†*[37] | IHS†*[35] | FAGLSUD |
| 1 | CO$_2$ (Kg) | Mean | 1705.58 | 1865.16 | 1735.73 | 1521.29 | 1550.47 | 1522.83 | 1530.44 | 1519.29 | 1539.77 | **1518.73** |
|   |   | Rank | 8 | 10 | 9 | 3 | 7 | 4 | 5 | 2 | 6 | **1** |
|   |   | SD | 36.02 | 46.93 | 62.13 | 6.22 | **0** | 8.98 | 8.31 | 0.39 | 12.25 | 5.53 |
| 2 | CO$_2$ (Kg) | Mean | 2045.61 | 2206.90 | 2038.34 | 1923.36 | 1946.65 | 1936.07 | 1924.07 | 2001.08 | 1945.55 | **1918.97** |
|   |   | Rank | 9 | 10 | 8 | 2 | 6 | 4 | 3 | 7 | 5 | **1** |
|   |   | SD | 23.41 | 45.79 | 44.97 | 6.39 | **0** | 17.17 | 4.54 | 9.89 | 19.69 | 5.53 |
| 3 | CO$_2$ (Kg) | Mean | 2680.26 | 2789.15 | 2505.33 | 2392.94 | 2399.99 | 2409.05 | 2435.96 | 2387.49 | 2410.77 | **2387.40** |
|   |   | Rank | 9 | 10 | 8 | 3 | 4 | 5 | 7 | 2 | 6 | **1** |
|   |   | SD | 95.10 | 74.53 | 56.04 | 12.68 | **0** | 17.07 | 34.93 | 11.82 | 11.81 | 11.71 |
| 4 | CO$_2$ (Kg) | Mean | 1586.21 | 1732.49 | 1618.00 | 1404.99 | 1403.73 | 1403.70 | 1410.50 | 1403.30 | 1421.76 | **1403.28** |
|   |   | Rank | 8 | 10 | 9 | 5 | 4 | 3 | 6 | 2 | 7 | **1** |
|   |   | SD | 37.22 | 46.12 | 55.96 | 4.68 | 5.13 | **4.39** | 6.59 | 4.91 | 10.99 | 4.85 |
| 5 | CO$_2$ (Kg) | Mean | 1912.39 | 2075.13 | 1924.86 | 1799.73 | 1856.41 | 1809.59 | 1798.51 | 1797.48 | 1812.96 | **1797.11** |
| | | Rank | 8 | 10 | 9 | 4 | 7 | 5 | 3 | 2 | 6 | **1** |
| | | SD | 24.57 | 44.73 | 43.35 | 4.98 | **0** | 15.07 | 4.07 | 6.32 | 18.19 | 5.96 |
| 6 | $CO_2$ (Kg) | Mean | 2571.06 | 2706.46 | 2437.73 | 2342.87 | 2366.00 | 2361.46 | 2355.67 | 2338.12 | 2363.05 | **2338.04** |
| | | Rank | 9 | 10 | 8 | 3 | 7 | 5 | 4 | 2 | 6 | **1** |
| | | SD | 71.17 | 64.02 | 53.91 | 5.53 | **0** | 14.69 | 20.15 | 5.27 | 16.56 | 5.39 |
| 7 | $CO_2$ (Kg) | Mean | 1476.87 | 1609.95 | 1485.25 | 1265.12 | 1273.20 | 1262.61 | 1270.66 | 1262.98 | 1281.42 | **1262.60** |
| | | Rank | 8 | 10 | 9 | 4 | 6 | 2 | 5 | 3 | 7 | **1** |
| | | SD | 41.15 | 48.75 | 59.25 | 3.96 | **0** | 3.10 | 7.12 | 3.29 | 9.42 | 3.09 |
| 8 | $CO_2$ (Kg) | Mean | 1793.83 | 1956.17 | 1810.07 | 1670.51 | 1669.86 | 1673.18 | 1668.44 | 1668.66 | 1677.83 | **1668.25** |
| | | Rank | 8 | 10 | 9 | 5 | 4 | 6 | 2 | 3 | 7 | **1** |
| | | SD | 29.01 | 44.95 | 47.28 | 4.59 | **0** | 9.67 | 2.96 | 7.81 | 11.82 | 3.75 |
| 9 | $CO_2$ (Kg) | Mean | 2612.23 | 2724.29 | 2456.10 | 2353.55 | 2358.57 | 2371.26 | 2359.45 | 2347.64 | 2376.88 | **2347.29** |
| | | Rank | 9 | 10 | 8 | 3 | 4 | 6 | 5 | 2 | 7 | **1** |
| | | SD | 115.91 | 66.34 | 62.72 | 8.62 | **0** | 15.40 | 10.98 | 3.32 | 17.20 | 3.07 |
| **Average rank** | | | 8.44 | 10 | 8.56 | 3.56 | 5.44 | 4.44 | 4.44 | 2.78 | 6.33 | **1** |
| **Overall rank** | | | 8 (7) | 10 (9) | 9 (8) | 3 | 6 (5) | 4.5 (4) | 4.5 (4) | 2 | 7 (6) | **1** |

### 7.3 Statistical significance analysis

Due to the non-deterministic nature of metaheuristics, we employ the Friedman test and the two-tailed Wilcoxon signed rank (TWSR) test [91] to determine whether the difference between our proposed approach and any of the other techniques is statistically significant in terms of the mean of three objectives—namely cost, weight, and $CO_2$. The relation between the TWSR test statistic value ($W_{stat}$) and critical value $W_{crit}$ can specify statistical significance on the mean objective values. If the obtained $W_{stat}$ value is more than the $W_{crit}$ at the significance level $\alpha=0.05$, then the null hypothesis $H_0$ is rejected, and the alternative hypothesis $H_1$ is accepted. Here, the null hypothesis states that the algorithms do not differ statistically, while the alternative hypothesis indicates a statistical difference between the algorithms. We individually analyzed the results of both example 1 and 2; each example includes three sets of objective values, so each set does not exhibit a particular distribution such as normal distribution or homogeneity of variance. According to the TWSR test table, in each set, the $W_{crit}$ for n = 9 case observations and $\alpha$ = 0.05 is 5. As already demonstrated in the previous section, in most of the case observations, FAGLSUD yields the lowest average rank followed by MSA, PSO, and LFBBO, and other methods. The results of the statistical analysis on paired comparisons between FAGLSUD and each RCC design optimization algorithms, given at the top right corner of each comparison method and indicated by the symbol † in Table 11 and Table 12, confirm that our proposed method offers statistically superior accuracy.

Table C 1 in Appendix C provides further details of our pairwise comparisons with the competitive methods (the second, third, and fourth-best overall ranked methods such as MSA, PSO, and LFBBO). We can see in the pairwise comparison of FAGLSUD vs. MSA that the $W_{stat}$

is 0 and thus lower than $W_{crit}$ = 5, signifying that there is a statistical difference between FAGLSUD and MSA in all sets of case observations. Similar trend is also observed for the pairwise comparison between FAGLSUD and LFBBO. Furthermore, the pairwise comparison of FAGLSUD vs. PSO explains that FAGLSUD provides significantly superior accuracy in 5 of the 6 sets of case observations; but they performed similarly in 1 set of case observations (the set of mean weight values in example 2), due to their similar characteristics such as global best-based update mechanism.

### 7.3 Conclusions

FAGLSUD is a novel optimization algorithm that mimics the hybrid evolution and swarm intelligence process of sub-populations, known as empires. Following its development in exploitation, exploration, and information sharing capabilities based on dynamic and rotational movement relations constituting fuzzy adaptive search operators (FAGLVA, FAUDVD, and FAEDELS), FAGLSUD was applied for the first time to the optimization of RCC walls, a constrained, twelve-dimensional optimization problem in structural engineering. Despite previously reported results found in the literature, no other study has focused on: (1) the dynamic analysis, which is considerably important particularly in earthquake-prone zones, or (2) an environmentally friendly design that contributes to the global warming issue.

To fill this gap, the present study applied FAGLSUD to identify the lowest cost, weight, and $CO_2$ emission designs of RCC walls based on nine different pseudo-static coefficients in various simulations. Furthermore, two example studies were conducted to evaluate both low and high values of the design parameters, such as the wall height and surcharge loading. Moreover, the seismic performance of each sample study was investigated against earthquake disasters.

It was found that increasing the wall height and surcharge loading results in designs with high cost, weights, and $CO_2$ emission. However, the impact of an earthquake is notably minor on a low-cost and low-$CO_2$ wall design with greater height and surcharge loading (example 2), where less changes in the objective values were observed for different loading cases ranging from 0 to 0.3. Contrarily, the impact of an earthquake was found to be greater on low-weight designs, as more sensitivity was observed towards the variation of earthquake coefficients. In example 2, the FAGLSUD lead to more optimal solutions than the other methods, indicating that more complexity in design does not affect the ability of the method to track the global optimal solution. In both examples, it was observed that vertical coefficient values ranging 0-0.3 and the horizontal coefficient values fixed at 0 or 0.15 yield lower final cost, weight, and $CO_2$ values. However, the trend reversed when the values of horizontal loading condition increased from 0 to 0.3. For the most severe surcharge loading case, where both horizontal and vertical coefficient values were fixed at 0.3, FAGLSUD performed the best.

We simulated two examples and analyzed three sets of observations for each example, including low-cost, low-weight and low-$CO_2$ designs. Each design was investigated against nine different pseudo-static coefficients denoted as loading cases 1 to 9. As we used metaheuristic optimization methods, which are non-deterministic, we reported the mean and standard deviation of final low-objective function values over 101 independent runs to measure the accuracy and the robustness of each optimization method against the random initial solutions. We then performed a statistical analysis using Friedman ranks followed by two-tailed Wilcoxon signed rank (TWSR). All results showed that, overall, the FAGLSUD was the best performing method; the MSA, PSO and LFBBO were ranked the second, third and fourth overall, respectively in most of the case observations. The TWSR confirmed that the proposed FAGLSUD offers statistically significantly superior accuracy compared to other RCC optimization methods. The only exception was the set of mean weight values in example 2, where the performance of PSO was similar to FAGLSUD, which is due to some common characteristics between the two methods (such as global best-based update mechanism).

We also evaluated three objective functions across 1000 iterations for all case studies. Some methods such as the MSA surpass FAGLSUD across 10-50% of the entire iteration history owing to their high exploitation capability. However, they do not offer superior accuracy at the end of convergence time. We concluded from our results that fast convergence tendency can significantly reduce the global search (exploration) capabilities of the algorithm and increases the likelihood of the search ending up at a local minimum. Further, and importantly, we have evaluated the stability of RCC design optimization results over 101 independent runs using SD criterion and observed that MSA and many other fast optimizers failed to offer superior stability in most of the case observations, clearly indicating that these methods suffer from sensitivity to initialization. Only LFBBO, in almost all case observations, offers superior stability in this respect, but it is not capable of handling the entire optimization process with satisfactory accuracy and efficiency (or convergence speed), as it shows higher objective values and slower convergence rates. On the other hand, FAGLSUD leverages high exploration capacity and despite not being among the best in terms of the convergence rate can improve the performance during the later iterations, with better stability and efficiency than many of the comparison methods. This can be attributed to its relatively excellent exploration, exploitation and information sharing capabilities, combined with appropriate parameter settings and operator selection due to the fuzzy adaptive universal diversity-based velocity divergence, fuzzy adaptive global learning-based velocity adaptation, fuzzy adaptive enhanced differential evolution-based local search policies and fuzzy adaptive operator selection. In addition, one may seed the initial population with feasible solutions to increase the computational efficiency in the early stages of optimization [79].

It should be noted that the compared algorithms were also successful in solving the optimization problem presented in this study, which can be owed to the best parameter settings. For instance, FPA was reported to adopt the Levy flight scaling factor of 0.5, offering the best convergence rate and efficiency compared to traditional optimizers like GA and PSO. However, the proposed FAGLSUD still outperformed several traditional and recently proposed algorithms, including ICA, DE, PSO, FPA, GWO, LFBBO, BBO, MSA, and IHS, yielding lower cost, weight, and $CO_2$ values with smaller variability (standard deviation) based on the analyses of 9 different seismic conditions and associated earthquake impact. Furthermore, the proposed method does not require parameter tuning to improve its performance, as is needed for other methods in literature.

The proposed FAGLSUD is an extended variant of the Fuzzy Adaptive Enhanced Imperialist Competitive Algorithm (FAEICA) [53]. Based on the FAGLSUD's promising results over twelve-dimensional RCC retaining wall designs, as well as the consistent behavior across various OPs with varying numbers of objectives, dimensions (from 1 to 30; low, multi, and high-dimensional OPs), and local optima demonstrated by similar fuzzy rule-based parameter adaptation and operator selection mechanisms [53], we can say that this study's findings have practical implications for a variety of constrained high-dimensional real-world optimization problems. This includes the optimum design of shallow foundation and slope stability analysis, which necessitates powerful exploration, exploitation, and search policies for information sharing. Consequently, with an increase in the number of design variables which leads to an expansion of the search space, adaptive velocity limit function (AVLF) and powerful search mechanisms (i.e., FAGLVA, FAUDVD, FAEDELS, and FAOS) designed for FAGLSUD (an extension variant of FAEICA [53]) are expected to provide effective search capability in high dimensions. The AVLF provides high exploration and exploitation, during the early and later stages of convergence, respectively, which improves the FAGLSUD's overall convergence behavior using less computational time. The FAGLVA, FAUDVD, FAEDELS, and FAOS contribute to exploitation, exploration and information sharing capacities of FAGLSUD and selection of the most prominent operators at each optimization stage, respectively. Furthermore, from a theoretical perspective, the proposed method can be further developed to tackle other complexity levels, including dynamism (moving optimum solutions), multi-objectivity (two or three objectives in conflict with each other, e.g., simultaneous improvement of accuracy and stability by a novel bio-objective optimization process), and many-objectivity (more than 3 objectives). The latter issues are common in real-world optimization problems, from the multi-objective optimization of engineering design problems to hybrid forecasting systems, the parameterization and computational costs should be addressed in addition to identifying the trade-offs between the contradictory objectives. Therefore, further studies are suggested to further improve the proposed method to overcome greater complexity levels and to mitigate other environmental impacts of RCC wall design that contribute to global warming.

**Declarations**

Funding: The first author acknowledges the support of an Australian Government Research Training Program scholarship to carry out this research.
Conflicts of interest: The authors declare that they have no conflict of interest.
Ethical approval: This article does not contain any studies with human participants or animals performed by any of the authors.
Informed consent: Not applicable.
Data availability: All data included in this study are available from the first author and can also be found in the manuscript.
Code availability: All code included in this study are available from the first author upon reasonable request.

**References**

1. Gandomi, A.H., et al., *Optimization of retaining wall design using evolutionary algorithms.* Structural and Multidisciplinary Optimization, 2017. **55**(3): p. 809-825.
2. Mergos, P.E. and F. Mantoglou, *Optimum design of reinforced concrete retaining walls with the flower pollination algorithm.* Structural and Multidisciplinary Optimization, 2020. **61**(2): p. 575-585.
3. Long, W., et al., *Solving high-dimensional global optimization problems using an improved sine cosine algorithm.* Expert Systems with Applications, 2019. **123**: p. 108-126.
4. Sarıbaş, A. and F. Erbatur, *Optimization and sensitivity of retaining structures.* Journal of Geotechnical Engineering, 1996. **122**(8): p. 649-656.
5. Basudhar, P., et al., *Cost optimization of reinforced earth walls.* Geotechnical and Geological Engineering, 2008. **26**(1): p. 1-12.
6. Sarma, K.C. and H. Adeli, *Cost optimization of concrete structures.* Journal of structural engineering, 1998. **124**(5): p. 570-578.

7. Wang, Y. and F.H. Kulhawy, *Economic design optimization of foundations.* Journal of geotechnical and geoenvironmental engineering, 2008. **134**(8): p. 1097-1105.
8. Hernández, S. and A.N. Fontán, *Practical applications of design optimization.* Vol. 3. 2002: Wit Press.
9. Fletcher, R., *Practical methods of optimization.* 2013: John Wiley & Sons.
10. Mourelatos, Z.P. and J. Liang. *An efficient unified approach for reliability and robustness in engineering design.* in *NSF Workshop on Reliable Engineering Computing.* 2004.
11. Sivakumar Babu, G. and B.M. Basha, *Optimum design of cantilever retaining walls using target reliability approach.* International journal of geomechanics, 2008. **8**(4): p. 240-252.
12. Weise, T., et al., *Why is optimization difficult?*, in *Nature-inspired algorithms for optimisation.* 2009, Springer. p. 1-50.
13. Kumar, M., A.J. Kulkarni, and S.C. Satapathy, *Socio evolution & learning optimization algorithm: A socio-inspired optimization methodology.* Future Generation Computer Systems, 2018. **81**: p. 252-272.
14. Eberhart, R. and J. Kennedy. *A new optimizer using particle swarm theory.* in *MHS'95. Proceedings of the Sixth International Symposium on Micro Machine and Human Science.* 1995. Ieee.
15. Yang, X.-S., *Engineering optimization: an introduction with metaheuristic applications.* 2010: John Wiley & Sons.
16. Dorigo, M., M. Birattari, and T. Stutzle, *Ant colony optimization.* IEEE computational intelligence magazine, 2006. **1**(4): p. 28-39.
17. Atashpaz-Gargari, E. and C. Lucas. *Imperialist competitive algorithm: an algorithm for optimization inspired by imperialistic competition.* in *2007 IEEE congress on evolutionary computation.* 2007. Ieee.
18. Yang, X.-S. and S. Deb. *Cuckoo search via Lévy flights.* in *2009 World congress on nature & biologically inspired computing (NaBIC).* 2009. Ieee.
19. Yang, X.-S., *Firefly algorithm.* Nature-inspired metaheuristic algorithms, 2008. **20**: p. 79-90.
20. Hasançebi, O. and S.K. Azad, *Adaptive dimensional search: a new metaheuristic algorithm for discrete truss sizing optimization.* Computers & Structures, 2015. **154**: p. 1-16.
21. Mirjalili, S., S.M. Mirjalili, and A. Lewis, *Grey wolf optimizer.* Advances in engineering software, 2014. **69**: p. 46-61.
22. Rechenberg, I., *Cybernetic solution path of an experimental problem.* Royal Aircraft Establishment Library Translation 1122, 1965.
23. Holland, J., *adaptation in natural and artificial systems, university of michigan press, ann arbor,".* Cité page, 1975. **100**.
24. Storn, R. *On the usage of differential evolution for function optimization.* in *Proceedings of North American Fuzzy Information Processing.* 1996. IEEE.
25. Kaveh, A. and M. Khayatazad, *A new meta-heuristic method: ray optimization.* Computers & structures, 2012. **112**: p. 283-294.
26. Civicioglu, P., *Backtracking search optimization algorithm for numerical optimization problems.* Applied Mathematics and computation, 2013. **219**(15): p. 8121-8144.
27. Erol, O.K. and I. Eksin, *A new optimization method: big bang–big crunch.* Advances in Engineering Software, 2006. **37**(2): p. 106-111.

28. Simon, D., *Biogeography-based optimization.* IEEE transactions on evolutionary computation, 2008. **12**(6): p. 702-713.
29. Yang, X.-S., *Firefly algorithm, Levy flights and global optimization*, in *Research and development in intelligent systems XXVI*. 2010, Springer. p. 209-218.
30. Gandomi, A., A. Kashani, and F. Zeighami, *Retaining wall optimization using interior search algorithm with different bound constraint handling.* International Journal for Numerical and Analytical Methods in Geomechanics, 2017. **41**(11): p. 1304-1331.
31. Eberhart, R.C., Y. Shi, and J. Kennedy, *Swarm intelligence*. 2001: Elsevier.
32. Geem, Z.W., J.H. Kim, and G.V. Loganathan, *A new heuristic optimization algorithm: harmony search.* simulation, 2001. **76**(2): p. 60-68.
33. Kirkpatrick, S., C.D. Gelatt, and M.P. Vecchi, *Optimization by simulated annealing.* science, 1983. **220**(4598): p. 671-680.
34. Azad, S.K., *Monitored convergence curve: a new framework for metaheuristic structural optimization algorithms.* Structural and Multidisciplinary Optimization, 2019. **60**(2): p. 481-499.
35. Kaveh, A. and M.A.A. Shakouri, *Harmony search based algorithms for the optimum cost design of reinforced concrete cantilever retaining walls.* International Journal of Civil Engineering, 2011. **9**(1): p. 1-8.
36. Mahdavi, M., M. Fesanghary, and E. Damangir, *An improved harmony search algorithm for solving optimization problems.* Applied mathematics and computation, 2007. **188**(2): p. 1567-1579.
37. Ceranic, B., C. Fryer, and R. Baines, *An application of simulated annealing to the optimum design of reinforced concrete retaining structures.* Computers & Structures, 2001. **79**(17): p. 1569-1581.
38. Aydogdu, I., *Cost optimization of reinforced concrete cantilever retaining walls under seismic loading using a biogeography-based optimization algorithm with Levy flights.* Engineering Optimization, 2017. **49**(3): p. 381-400.
39. Kashani, A.R., et al., *Metaheuristics in civil engineering: A review.* 1, 2020. **1**(1): p. 019.
40. Kashani, A.R., et al., *Particle Swarm Optimization Variants for Solving Geotechnical Problems: Review and Comparative Analysis.* Archives of Computational Methods in Engineering, 2020: p. 1-57.
41. Kalemci, E.N., et al. *Design of reinforced concrete cantilever retaining wall using Grey wolf optimization algorithm.* in *Structures*. 2020. Elsevier.
42. Committee, A. *Building code requirements for structural concrete (ACI 318-05) and commentary (ACI 318R-05)*. 2005. American Concrete Institute.
43. STANDARD, B., *Eurocode 7: Geotechnical design—*. 2004, 2004.
44. Al Atik, L. and N. Sitar, *Seismic earth pressures on cantilever retaining structures.* Journal of Geotechnical and Geoenvironmental engineering, 2010. **136**(10): p. 1324-1333.
45. Candia, G., R.G. Mikola, and N. Sitar, *Seismic response of retaining walls with cohesive backfill: Centrifuge model studies.* Soil Dynamics and Earthquake Engineering, 2016. **90**: p. 411-419.
46. Geraili Mikola, R., G. Candia, and N. Sitar, *Seismic earth pressures on retaining structures and basement walls in cohesionless soils.* Journal of Geotechnical and Geoenvironmental Engineering, 2016. **142**(10): p. 04016047.

47. Kaveh, A., S. Talatahari, and R. Sheikholeslami. *Optimum seismic design of gravity retaining walls using the heuristic big bang-big crunch algorithm*. in *Proceedings of the Second International Conference on Soft Computing Technology in Civil, Structural and Environmental Engineering, Civil-Comp Press, Stirlingshire, Scotland, Paper*. 2011.
48. Kaveh, A., M. Kalateh-Ahani, and M. Fahimi-Farzam, *Constructability optimal design of reinforced concrete retaining walls using a multi-objective genetic algorithm.* Structural Engineering and Mechanics, 2013. **47**(2): p. 227-245.
49. Gandomi, A.H. and A.R. Kashani, *Automating pseudo-static analysis of concrete cantilever retaining wall using evolutionary algorithms.* Measurement, 2018. **115**: p. 104-124.
50. Mergos, P.E., *Contribution to sustainable seismic design of reinforced concrete members through embodied CO2 emissions optimization.* Structural Concrete, 2018. **19**(2): p. 454-462.
51. Georgopoulos, C., & Minson, A., *Sustainable concrete solutions*. 2014: John Wiley & Sons.
52. Olivier, J.G., J.A. Peters, and G. Janssens-Maenhout, *Trends in global CO2 emissions 2012 report.* 2012.
53. Keivanian, F. and R. Chiong, *A Novel Hybrid Fuzzy–Metaheuristic Approach for Multimodal Single and Multi-Objective Optimization Problems.* Expert Systems with Applications, 2021: p. 116199.
54. Hansbo, S., *Foundation engineering*. 1994: Newnes.
55. Camp, C.V. and A. Akin, *Design of retaining walls using big bang–big crunch optimization.* Journal of Structural Engineering, 2012. **138**(3): p. 438-448.
56. Okabe, S., *General theory on earth pressure and seismic stability of retaining wall and dam.* Proc. Civil Engrg. Soc., Japan, 1924. **10**(6): p. 1277-1323.
57. Mononobe, N. and H. Matsuo. *On the determination of earth pressures during earthquakes" volume 9, Tokyo, 1929*. in *World Engineering Congress*.
58. Kramer, S.L., *Geotechnical earthquake engineering*. 1996: Pearson Education India.
59. Das, B.M., *Principles of foundation engineering*. 2015: Cengage learning.
60. Coulomb, C.A., *Essai sur une application des regles de maximis et minimis a quelques problemes de statique relatifs a l'architecture (essay on maximums and minimums of rules to some static problems relating to architecture).* 1973.
61. Seed, H. *Design of earth retaining structures for dynamic loads*. in *ASCE Specialty Conf.-Lateral Stress in the Ground and Design of Earth Retaining Structures, 1970*. 1970.
62. AASHTO, L., *Standard specifications for highway bridges.* Officials, Seventeenth Edition, American Association of State Highway and Transportation Washington, DC, 2002.
63. Technology, C.I.o.C., *BEDEC PR/PCT ITEC materials database*, T. CIoC, Editor. 2009: Barcelona.
64. Yepes, V., et al., *CO2-optimization design of reinforced concrete retaining walls based on a VNS-threshold acceptance strategy.* Journal of Computing in Civil Engineering, 2012. **26**(3): p. 378-386.
65. Zadeh, L.A., *Fuzzy sets as a basis for a theory of possibility.* Fuzzy sets and systems, 1978. **1**(1): p. 3-28.

66. Zimmermann, H.-J., *Fuzzy set theory—and its applications*. 2011: Springer Science & Business Media.
67. Antão, R., *Type-2 Fuzzy Logic: Uncertain Systems' Modeling and Control*. 2017: Springer.
68. Zadeh, L.A., *The concept of a linguistic variable and its application to approximate reasoning—I.* Information sciences, 1975. **8**(3): p. 199-249.
69. Mamdani, E.H. *Application of fuzzy algorithms for control of simple dynamic plant*. in *Proceedings of the institution of electrical engineers*. 1974. IET.
70. Kerr-Wilson, J. and W. Pedrycz, *Generating a hierarchical fuzzy rule-based model.* Fuzzy Sets and Systems, 2020. **381**: p. 124-139.
71. Drobics, M. and J. Botzheim, *Optimization of fuzzy rule sets using a bacterial evolutionary algorithm.* Mathware & Soft Computing, 2008. **15**(1): p. 21-40.
72. Keivanian, F., N. Mehrshad, and S.-H. Zahiri, *Optimum layout of multiplexer with minimal average power based on IWO, fuzzy-IWO, GA, and fuzzy GA.* ACSIJ Adv Comput Sci Int J, 2014. **3**(5): p. 132-139.
73. Keivanian, F., N. Mehrshad, and A. Bijari, *Multi-objective optimization of MOSFETs channel widths and supply voltage in the proposed dual edge-triggered static D flip-flop with minimum average power and delay by using fuzzy non-dominated sorting genetic algorithm-II.* SpringerPlus, 2016. **5**(1): p. 1-15.
74. Keivanian, F., *Minimization of average power consumption in 3 stage CMOS ring oscillator based on MSFLA, fuzzy-MSFLA, GA, and fuzzy-GA.* International Journal of Computer Applications, 2014. **104**(16).
75. Keivanian, F., R. Chiong, and Z. Hu. *A Fuzzy Adaptive Binary Global Learning Colonization-MLP model for Body Fat Prediction*. in *2019 3rd International Conference on Bio-engineering for Smart Technologies (BioSMART)*. 2019. IEEE.
76. Keivanian, F. and N. Mehrshad. *Intelligent feature subset selection with unspecified number for body fat prediction based on binary-GA and Fuzzy-Binary-GA*. in *2015 2nd International Conference on Pattern Recognition and Image Analysis (IPRIA)*. 2015. IEEE.
77. Keivanian, F., A. Yekta Awal, and N. Mehrshad, *Optimization of JK Flip Flop Layout with minimal average power of consumption based on ACOR, fuzzy-ACOR, GA, and fuzzy-GA.* J Math Comput Sci, 2014. **14**(1): p. 1-15.
78. Santiago, A., et al., *A novel multi-objective evolutionary algorithm with fuzzy logic based adaptive selection of operators: FAME.* Information Sciences, 2019. **471**: p. 233-251.
79. Kazemzadeh Azad, S., *Seeding the initial population with feasible solutions in metaheuristic optimization of steel trusses.* Engineering Optimization, 2018. **50**(1): p. 89-105.
80. Al Khaled, A. and S. Hosseini, *Fuzzy adaptive imperialist competitive algorithm for global optimization.* Neural Computing and Applications, 2015. **26**(4): p. 813-825.
81. Azad, S.K. and O. Hasançebi, *Upper bound strategy for metaheuristic based design optimization of steel frames.* Advances in Engineering Software, 2013. **57**: p. 19-32.
82. Hasançebi, O., S. Çarbas, and M. Saka, *A reformulation of the ant colony optimization algorithm for large scale structural optimization.* 2011.
83. Hancer, E., *Differential evolution for feature selection: a fuzzy wrapper–filter approach.* Soft Computing, 2019. **23**(13): p. 5233-5248.

84. Liang, J.J., et al., *Comprehensive learning particle swarm optimizer for global optimization of multimodal functions.* IEEE transactions on evolutionary computation, 2006. **10**(3): p. 281-295.
85. Hadi, A.A., A.W. Mohamed, and K.M. Jambi, *LSHADE-SPA memetic framework for solving large-scale optimization problems.* Complex & Intelligent Systems, 2019. **5**(1): p. 25-40.
86. Storn, R. and K. Price, *Differential evolution–a simple and efficient heuristic for global optimization over continuous spaces.* Journal of global optimization, 1997. **11**(4): p. 341-359.
87. Masoudi, E., H. Holling, and W.K. Wong, *Application of imperialist competitive algorithm to find minimax and standardized maximin optimal designs.* Computational Statistics & Data Analysis, 2017. **113**: p. 330-345.
88. Ji, X., et al., *An efficient imperialist competitive algorithm for solving the QFD decision problem.* Mathematical Problems in Engineering, 2016. **2016**.
89. Yepes, V., et al., *A parametric study of optimum earth-retaining walls by simulated annealing.* Engineering Structures, 2008. **30**(3): p. 821-830.
90. Medina, J.R., *Estimation of incident and reflected waves using simulated annealing.* Journal of waterway, port, coastal, and ocean engineering, 2001. **127**(4): p. 213-221.
91. Derrac, J., et al., *A practical tutorial on the use of nonparametric statistical tests as a methodology for comparing evolutionary and swarm intelligence algorithms.* Swarm and Evolutionary Computation, 2011. **1**(1): p. 3-18.

**Appendix A**

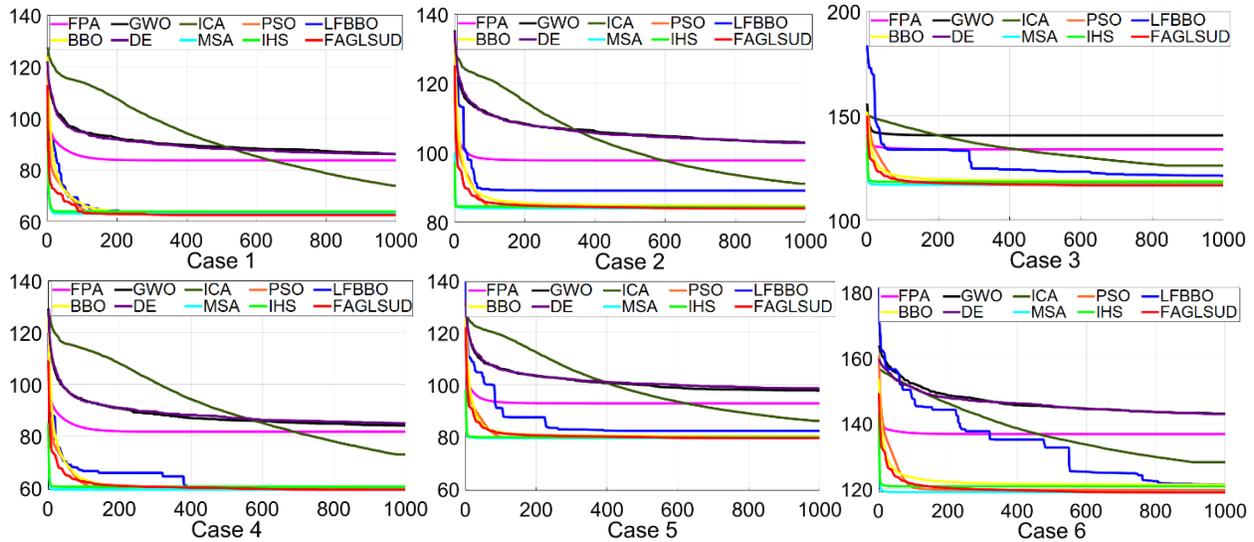

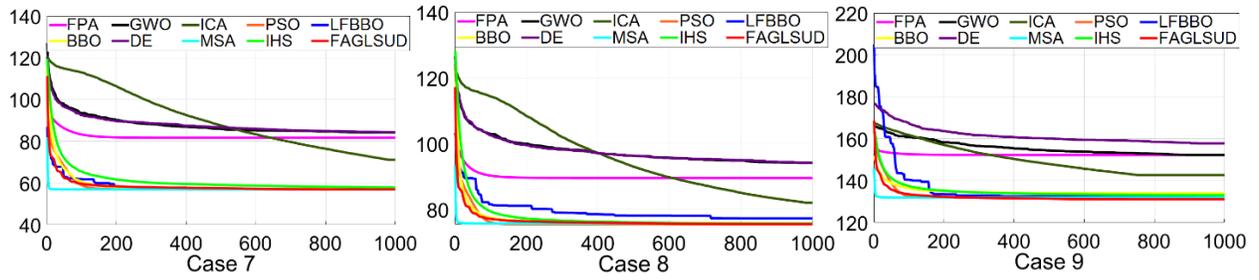

Figure A 1: Convergence rate plots based on average cost histories of example 1 under 9 seismic analyses (Cases 1 to 9)

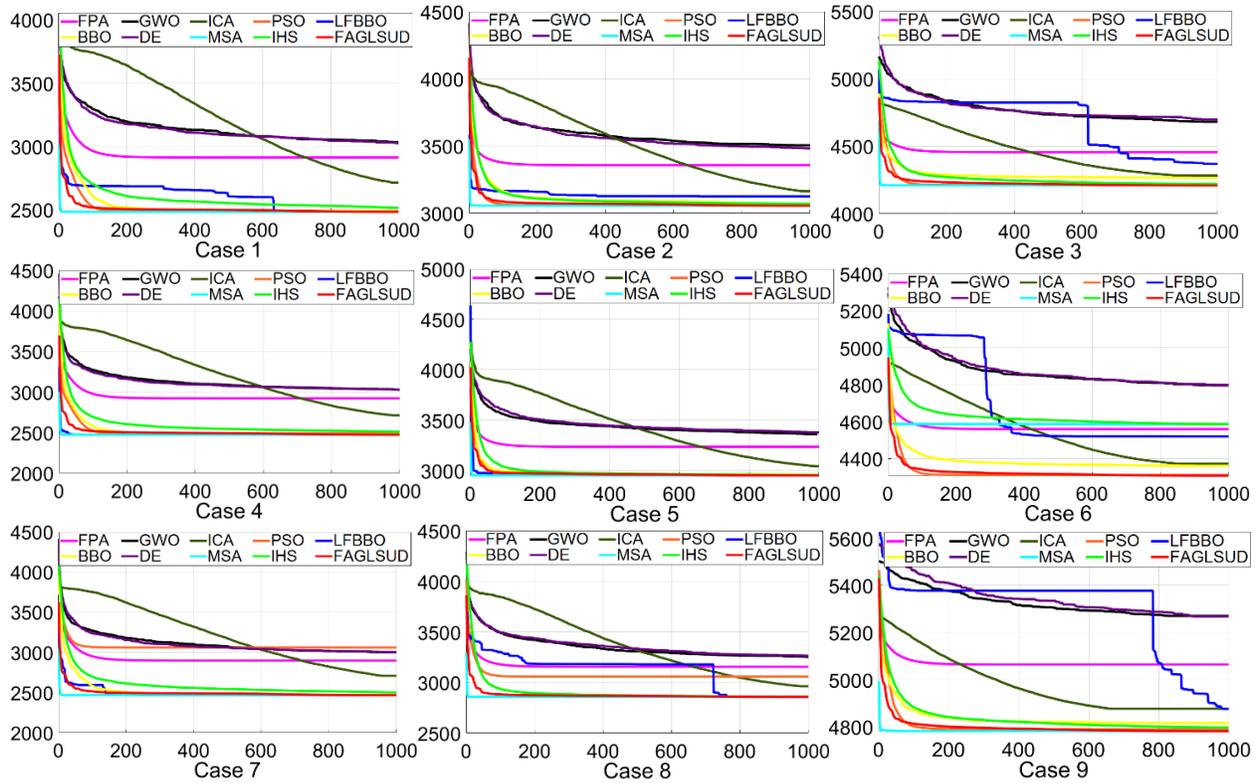

Figure A 2: Convergence rate plots based on average weight histories of example 1 under 9 seismic analyses (Cases 1 to 9)

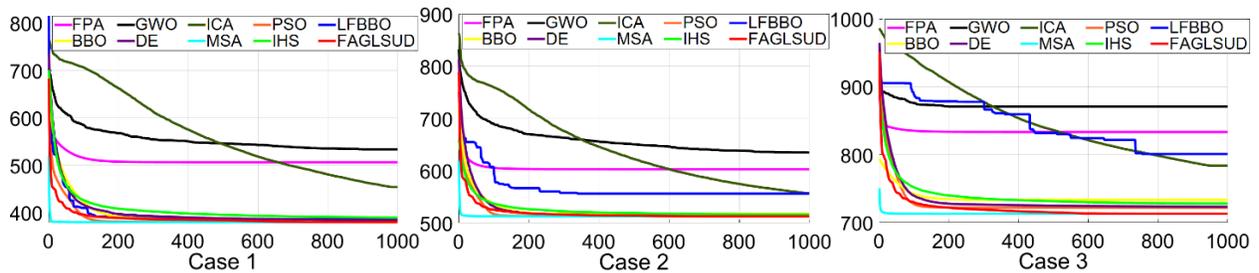

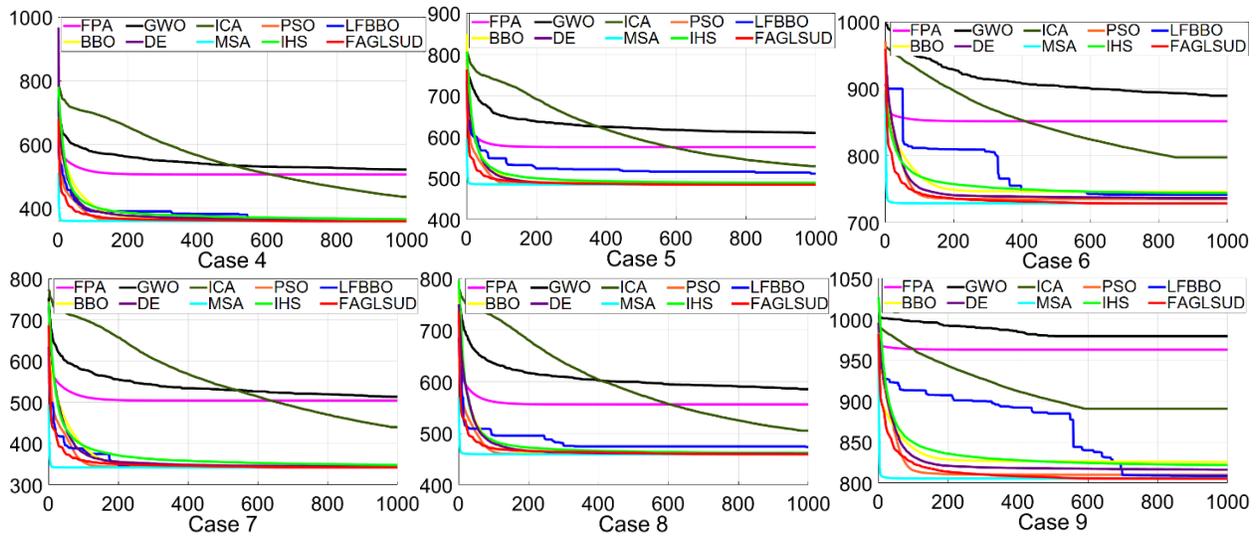

Figure A 3: Convergence rate plots based on average $CO_2$ histories of example 1 under 9 seismic analyses (Cases 1 to 9)

Table A 1: Final low best-cost retaining wall design variables along with their associated cost objective values for example 1 under 9 seismic analyses achieved by FAGLSUD

| Case | X1 | X2 | X3 | X4 | X5 | X6 | X7 | X8 | R1 | R2 | R3 | R4 | Cost |
|---|---|---|---|---|---|---|---|---|---|---|---|---|---|
| 1 | 1.51 | 0.78 | 0.20 | 0.20 | 0.27 | 1.31 | 0.20 | 0.20 | 28.03 | 17.98 | 17.96 | 7.37 | 62.33 |
| 2 | 2.09 | 0.78 | 0.27 | 0.20 | 0.27 | 1.39 | 0.20 | 0.20 | 45.29 | 14.09 | 14.47 | 7.51 | 83.42 |
| 3 | 2.86 | 0.78 | 0.33 | 0.20 | 0.27 | 2.17 | 0.20 | 0.20 | 82.67 | 14.60 | 14.28 | 7.81 | 115.97 |
| 4 | 1.51 | 0.78 | 0.20 | 0.20 | 0.27 | 1.31 | 0.20 | 0.20 | 14.72 | 14.12 | 14.18 | 7.84 | 59.27 |
| 5 | 2.00 | 0.78 | 0.26 | 0.20 | 0.27 | 1.43 | 0.20 | 0.20 | 37.56 | 14.19 | 14.21 | 7.15 | 79.00 |
| 6 | 2.93 | 0.78 | 0.33 | 0.20 | 0.27 | 1.38 | 0.20 | 0.20 | 82.90 | 14.44 | 14.06 | 7.46 | 118.38 |
| 7 | 1.51 | 0.78 | 0.20 | 0.20 | 0.27 | 1.31 | 0.20 | 0.20 | 6.67 | 14.71 | 14.04 | 7.18 | 56.88 |
| 8 | 1.91 | 0.78 | 0.24 | 0.20 | 0.27 | 1.31 | 0.20 | 0.20 | 33.02 | 14.92 | 14.07 | 7.97 | 74.88 |
| 9 | 3.17 | 0.78 | 0.32 | 0.20 | 0.27 | 1.31 | 0.20 | 0.20 | 102.56 | 14.42 | 14.57 | 7.04 | 130.83 |

Table A 2: Final low best-weight retaining wall design variables along with their associated weight objective values for example 1 under 9 seismic analyses achieved by FAGLSUD

| Case | X1 | X2 | X3 | X4 | X5 | X6 | X7 | X8 | R1 | R2 | R3 | R4 | Weight |
|---|---|---|---|---|---|---|---|---|---|---|---|---|---|
| 1 | 1.51 | 0.78 | 0.20 | 0.20 | 0.27 | 1.31 | 0.20 | 0.20 | 29.90 | 14.06 | 14.26 | 7.85 | 2480.95 |
| 2 | 2.06 | 0.78 | 0.20 | 0.20 | 0.27 | 1.45 | 0.33 | 0.33 | 92.96 | 14.41 | 14.91 | 23.63 | 3054.37 |
| 3 | 2.81 | 0.78 | 0.25 | 0.20 | 0.27 | 1.33 | 0.33 | 0.33 | 129.79 | 14.30 | 14.35 | 23.68 | 4206.79 |
| 4 | 1.51 | 0.78 | 0.20 | 0.20 | 0.27 | 1.31 | 0.20 | 0.20 | 14.57 | 14.84 | 14.89 | 7.38 | 2471.23 |
| 5 | 1.97 | 0.78 | 0.20 | 0.20 | 0.27 | 1.52 | 0.33 | 0.33 | 72.35 | 14.84 | 14.52 | 23.80 | 2950.26 |
| 6 | 2.88 | 0.78 | 0.25 | 0.20 | 0.27 | 1.31 | 0.20 | 0.20 | 129.26 | 14.12 | 14.23 | 7.16 | 4306.70 |
| 7 | 1.51 | 0.78 | 0.20 | 0.20 | 0.27 | 1.31 | 0.20 | 0.20 | 6.95 | 14.89 | 14.04 | 7.74 | 2465.25 |
| 8 | 1.89 | 0.78 | 0.20 | 0.20 | 0.27 | 1.31 | 0.20 | 0.20 | 56.80 | 14.18 | 14.19 | 7.63 | 2857.46 |
| 9 | 3.14 | 0.78 | 0.28 | 0.20 | 0.27 | 1.49 | 0.20 | 0.20 | 130.14 | 14.80 | 14.32 | 7.37 | 4779.02 |

Table A 3: Final low best-$CO_2$ retaining wall design variables along with their associated $CO_2$ objective values for example 1 under 9 seismic analyses achieved by FAGLSUD

| Case | X1 | X2 | X3 | X4 | X5 | X6 | X7 | X8 | R1 | R2 | R3 | R4 | $CO_2$ |
|---|---|---|---|---|---|---|---|---|---|---|---|---|---|
| 1 | 1.51 | 0.78 | 0.24 | 0.20 | 0.27 | 1.31 | 0.20 | 0.20 | 17.02 | 14.76 | 14.93 | 7.10 | 378.60 |
| 2 | 2.10 | 0.78 | 0.29 | 0.20 | 0.27 | 1.54 | 0.20 | 0.20 | 37.99 | 14.51 | 14.11 | 7.61 | 508.52 |
| 3 | 2.86 | 0.78 | 0.33 | 0.20 | 0.27 | 1.31 | 0.20 | 0.20 | 82.10 | 14.63 | 14.70 | 7.40 | 710.58 |
| 4 | 1.51 | 0.78 | 0.21 | 0.20 | 0.27 | 1.31 | 0.20 | 0.20 | 12.15 | 14.15 | 14.41 | 7.17 | 358.70 |
| 5 | 2.01 | 0.78 | 0.27 | 0.20 | 0.27 | 1.59 | 0.20 | 0.20 | 33.98 | 14.92 | 14.70 | 7.02 | 481.21 |
| 6 | 2.93 | 0.78 | 0.33 | 0.20 | 0.27 | 1.31 | 0.20 | 0.20 | 82.84 | 14.46 | 14.16 | 7.88 | 724.84 |
| 7 | 1.51 | 0.78 | 0.20 | 0.20 | 0.27 | 1.31 | 0.20 | 0.20 | 6.61 | 14.06 | 14.21 | 7.97 | 342.09 |
| 8 | 1.92 | 0.78 | 0.25 | 0.20 | 0.27 | 1.49 | 0.20 | 0.20 | 30.15 | 14.08 | 14.15 | 7.17 | 455.81 |
| 9 | 3.17 | 0.78 | 0.32 | 0.20 | 0.27 | 2.60 | 0.20 | 0.20 | 102.49 | 14.74 | 14.81 | 7.31 | 804.56 |

## Appendix B

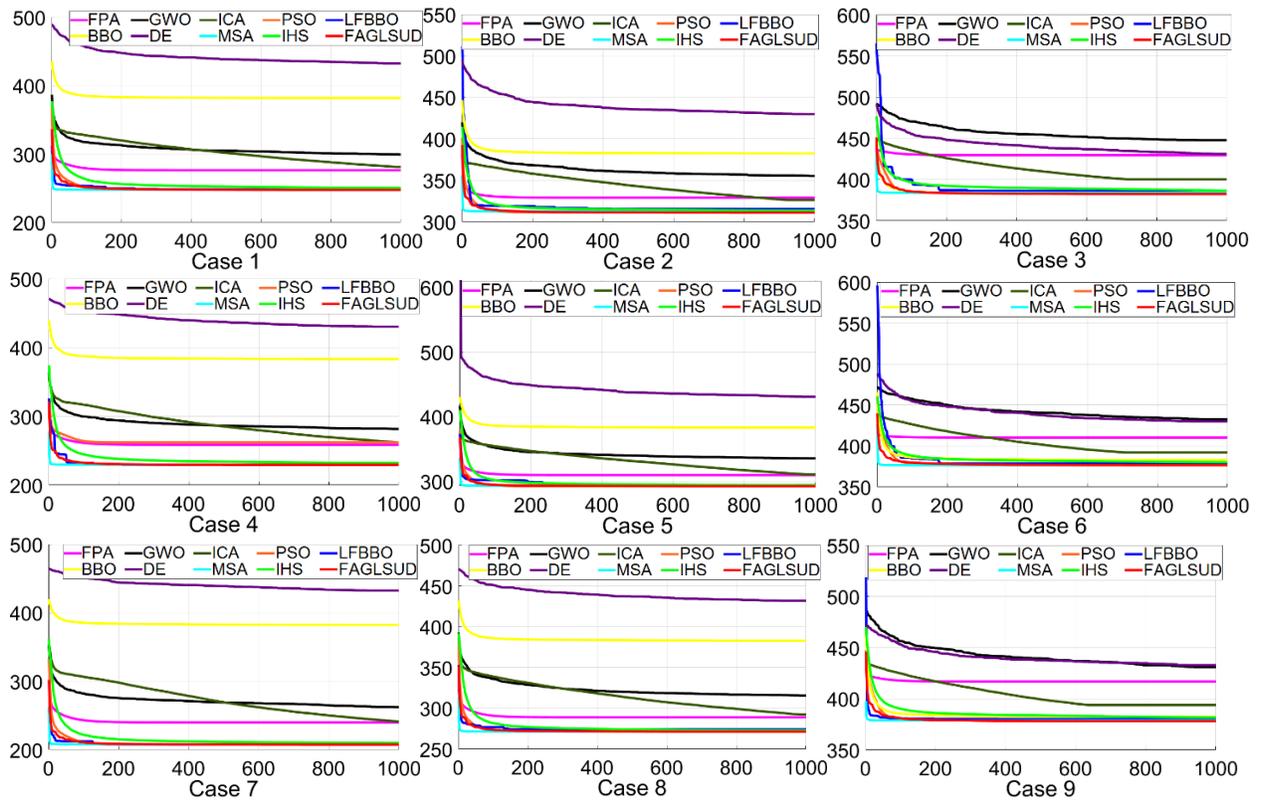

Figure B 1: Convergence rate plots based on average cost histories of example 2 under 9 seismic analyses (Cases 1 to 9)

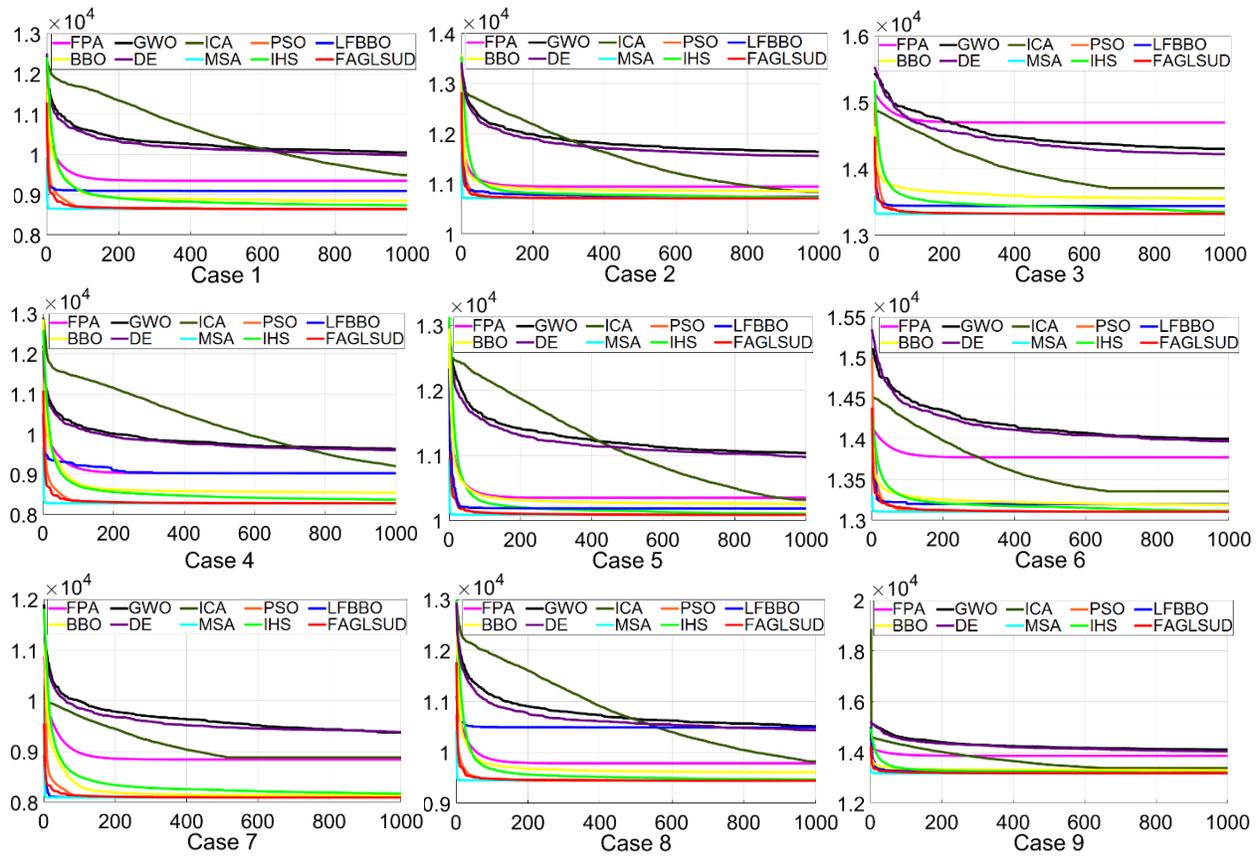

Figure B 2: Convergence rate plots based on average weight histories of example 2 under 9 seismic analyses (Cases 1 to 9)

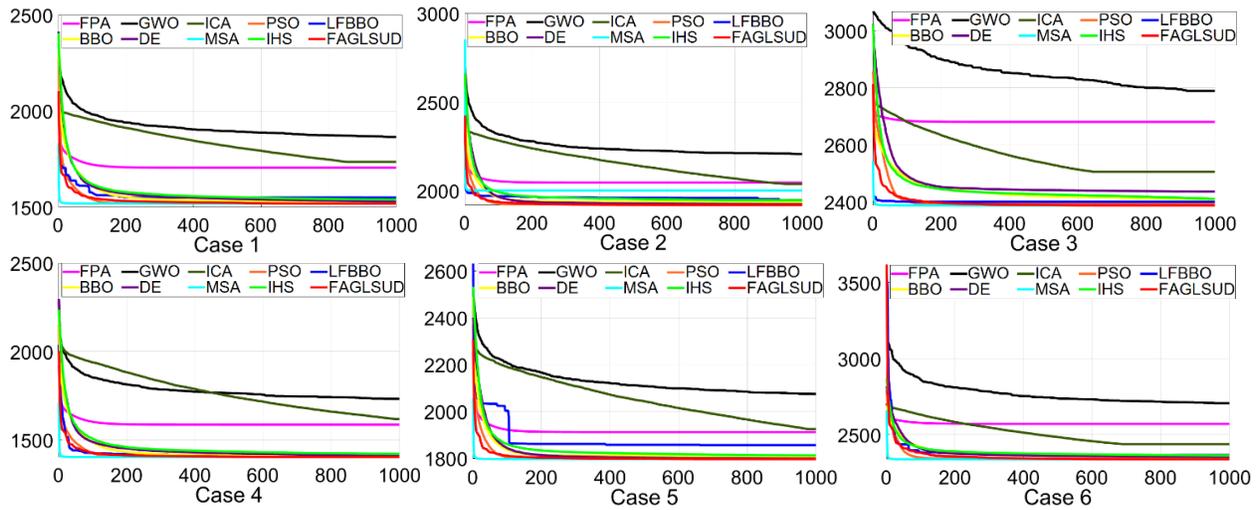

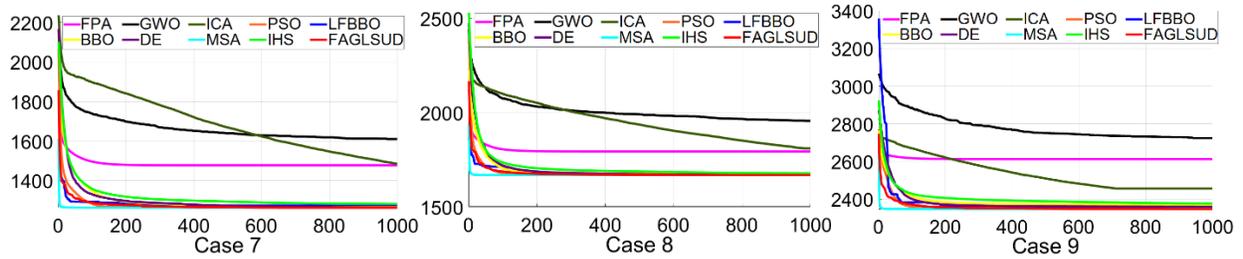

Figure B 3: Convergence rate plots based on average $CO_2$ histories of Example 2 under 9 seismic analyses (Cases 1 to 9)

Table B 1: Final low best-cost retaining wall design variables along with their associated cost objective values for example 2 under 9 seismic analyses achieved by FAGLSUD

| Case | X1 | X2 | X3 | X4 | X5 | X6 | X7 | X8 | R1 | R2 | R3 | R4 | Cost |
|---|---|---|---|---|---|---|---|---|---|---|---|---|---|
| 1 | 2.90 | 0.87 | 0.47 | 0.30 | 0.54 | 2.60 | 0.30 | 0.30 | 97.82 | 56.11 | 56.04 | 20.19 | 246.78 |
| 2 | 3.89 | 1.22 | 0.59 | 0.30 | 0.54 | 2.60 | 0.32 | 0.30 | 129.16 | 56.87 | 56.60 | 20.29 | 310.53 |
| 3 | 4.92 | 1.55 | 0.66 | 0.30 | 0.54 | 4.40 | 0.32 | 0.30 | 166.01 | 56.00 | 56.04 | 20.66 | 382.98 |
| 4 | 2.90 | 0.87 | 0.41 | 0.30 | 0.54 | 2.60 | 0.30 | 0.30 | 77.72 | 56.57 | 56.07 | 20.17 | 228.79 |
| 5 | 3.67 | 1.09 | 0.57 | 0.30 | 0.54 | 2.90 | 0.30 | 0.30 | 114.99 | 56.22 | 56.97 | 20.39 | 291.23 |
| 6 | 4.91 | 1.47 | 0.66 | 0.30 | 0.54 | 3.84 | 0.30 | 0.30 | 159.54 | 57.00 | 56.37 | 20.07 | 376.07 |
| 7 | 2.90 | 1.56 | 0.33 | 0.30 | 0.54 | 2.60 | 0.30 | 0.30 | 51.81 | 56.90 | 56.77 | 20.87 | 206.93 |
| 8 | 3.40 | 0.94 | 0.49 | 0.30 | 0.54 | 2.60 | 0.30 | 0.30 | 111.06 | 56.66 | 56.59 | 20.95 | 270.63 |
| 9 | 4.99 | 1.43 | 0.65 | 0.30 | 0.54 | 2.60 | 0.30 | 0.30 | 159.78 | 56.01 | 56.00 | 20.23 | 377.77 |

Table B 2: Final low best-weight retaining wall design variables along with their associated weight objective values for example 2 under 9 seismic analyses achieved by FAGLSUD

| Case | X1 | X2 | X3 | X4 | X5 | X6 | X7 | X8 | R1 | R2 | R3 | R4 | Weight |
|---|---|---|---|---|---|---|---|---|---|---|---|---|---|
| 1 | 2.90 | 1.02 | 0.36 | 0.30 | 0.54 | 2.60 | 0.30 | 0.30 | 139.21 | 56.90 | 56.64 | 20.20 | 8637.37 |
| 2 | 3.83 | 1.25 | 0.48 | 0.30 | 0.54 | 2.60 | 0.67 | 0.67 | 157.46 | 56.73 | 56.07 | 77.70 | 10706.23 |
| 3 | 4.91 | 1.56 | 0.65 | 0.30 | 0.54 | 2.60 | 0.67 | 0.67 | 170.11 | 56.28 | 56.11 | 77.66 | 13322.39 |
| 4 | 2.90 | 1.55 | 0.32 | 0.30 | 0.54 | 2.60 | 0.30 | 0.30 | 118.44 | 56.02 | 56.59 | 20.95 | 8274.76 |
| 5 | 3.61 | 1.07 | 0.43 | 0.30 | 0.54 | 3.17 | 0.30 | 0.30 | 152.05 | 56.53 | 56.75 | 20.33 | 10080.21 |
| 6 | 4.89 | 1.46 | 0.62 | 0.30 | 0.54 | 2.60 | 0.30 | 0.30 | 170.95 | 56.80 | 56.84 | 20.84 | 13101.59 |
| 7 | 2.90 | 1.02 | 0.30 | 0.30 | 0.54 | 2.60 | 0.30 | 0.30 | 61.13 | 56.76 | 56.66 | 20.01 | 8093.10 |
| 8 | 3.35 | 0.94 | 0.39 | 0.30 | 0.54 | 2.60 | 0.30 | 0.30 | 144.65 | 56.38 | 56.11 | 20.55 | 9439.61 |
| 9 | 4.97 | 1.42 | 0.62 | 0.30 | 0.54 | 2.71 | 0.30 | 0.30 | 170.88 | 56.97 | 56.55 | 20.07 | 13176.35 |

Table B 3: Final low best-$CO_2$ retaining wall design variables along with their associated $CO_2$ objective values for example 2 under 9 seismic analyses achieved by FAGLSUD

| Case | X1 | X2 | X3 | X4 | X5 | X6 | X7 | X8 | R1 | R2 | R3 | R4 | $CO_2$ |
|---|---|---|---|---|---|---|---|---|---|---|---|---|---|
| 1 | 2.90 | 0.87 | 0.51 | 0.30 | 0.54 | 2.60 | 0.30 | 0.30 | 90.02 | 56.01 | 56.69 | 20.21 | 1515.09 |
| 2 | 3.90 | 1.23 | 0.64 | 0.30 | 0.54 | 2.92 | 0.32 | 0.30 | 122.97 | 56.94 | 56.55 | 20.08 | 1913.69 |
| 3 | 4.92 | 1.54 | 0.66 | 0.30 | 0.54 | 3.91 | 0.32 | 0.30 | 166.78 | 56.83 | 56.07 | 20.81 | 2383.29 |
| 4 | 2.90 | 1.54 | 0.44 | 0.30 | 0.54 | 2.60 | 0.30 | 0.30 | 71.21 | 56.45 | 56.86 | 20.03 | 1400.15 |
| 5 | 3.67 | 1.09 | 0.59 | 0.30 | 0.54 | 2.60 | 0.30 | 0.30 | 111.56 | 56.14 | 56.17 | 20.36 | 1791.28 |
| 6 | 4.89 | 1.46 | 0.62 | 0.30 | 0.54 | 2.60 | 0.30 | 0.30 | 170.95 | 56.80 | 56.84 | 20.84 | 2336.12 |
| 7 | 2.90 | 1.05 | 0.34 | 0.30 | 0.54 | 2.60 | 0.30 | 0.30 | 47.35 | 56.44 | 56.76 | 20.12 | 1260.95 |
| 8 | 3.42 | 0.95 | 0.54 | 0.30 | 0.54 | 2.60 | 0.30 | 0.30 | 97.10 | 56.29 | 56.27 | 20.12 | 1663.48 |
| 9 | 4.99 | 1.43 | 0.65 | 0.30 | 0.54 | 2.60 | 0.30 | 0.30 | 159.73 | 56.15 | 56.83 | 20.90 | 2346.62 |

## Appendix C

Table C 1: Pairwise comparisons of FAGLSUD vs. the second, third, and fourth-based overall ranked methods such as MSA, PSO and LFBBO, with nine case observations $n = 9$, $\alpha = 0.05$, and $W_{crit} = 5$ in each set

| Example 1 | Mean Cost | Observations | FAGLSUD | MSA | Differences | Absolute differences | Rank | $W_{stat}$ |
|---|---|---|---|---|---|---|---|---|
| Example 1 | Mean Cost | Case 1 | 62.45 | 62.50 | -0.05 | 0.05 | 3.50 | 0 |
| | | Case 2 | 83.93 | 83.98 | -0.05 | 0.05 | 3.50 | ($<W_{crit}$) |
| | | Case 3 | 116.71 | 116.97 | -0.26 | 0.26 | 8 | |
| | | Case 4 | 59.35 | 59.38 | -0.03 | 0.03 | 2 | |
| | | Case 5 | 79.51 | 79.62 | -0.11 | 0.11 | 6 | |
| | | Case 6 | 118.90 | 119.00 | -0.10 | 0.10 | 5 | |
| | | Case 7 | 56.88 | 56.89 | -0.01 | 0.01 | 1 | |
| | | Case 8 | 75.44 | 75.67 | -0.23 | 0.23 | 7 | |
| | | Case 9 | 131.06 | 131.89 | -0.83 | 0.83 | 9 | |
| | Mean Weight | Case 1 | 2481.87 | 2481.97 | -0.10 | 0.10 | 3 | 0 |
| | | Case 2 | 3055.33 | 3055.42 | -0.09 | 0.09 | 2 | ($<W_{crit}$) |
| | | Case 3 | 4208.16 | 4209 | -0.84 | 0.84 | 7 | |
| | | Case 4 | 2472.46 | 2472.53 | -0.07 | 0.07 | 1 | |
| | | Case 5 | 2950.91 | 2951.20 | -0.29 | 0.29 | 4 | |
| | | Case 6 | 4308.41 | 4586.98 | -278.57 | 278.57 | 9 | |
| | | Case 7 | 2465.38 | 2466.45 | -1.07 | 1.07 | 8 | |
| | | Case 8 | 2858.19 | 2858.70 | -0.51 | 0.51 | 5 | |
| | | Case 9 | 4781.25 | 4782.02 | -0.77 | 0.77 | 6 | |
| | Mean $CO_2$ | Case 1 | 379.97 | 379.99 | -0.02 | 0.02 | 2.50 | 0 |
| | | Case 2 | 512.66 | 512.80 | -0.14 | 0.14 | 5 | ($<W_{crit}$) |
| | | Case 3 | 712.77 | 712.79 | -0.02 | 0.02 | 2.50 | |
| | | Case 4 | 359.07 | 359.80 | -0.73 | 0.73 | 8 | |
| | | Case 5 | 484.01 | 484.99 | -0.98 | 0.98 | 9 | |
| | | Case 6 | 728.47 | 728.84 | -0.37 | 0.37 | 7 | |
| | | Case 7 | 342.11 | 342.19 | -0.08 | 0.08 | 4 | |
| | | Case 8 | 459.37 | 459.38 | -0.01 | 0.01 | 1 | |
| | | Case 9 | 805.37 | 805.64 | -0.27 | 0.27 | 6 | |
| Example 2 | Mean Cost | Case 1 | 247.39 | 247.78 | -0.39 | 0.39 | 4.50 | 0 |
| | | Case 2 | 311.00 | 312.75 | -1.75 | 1.75 | 9 | ($<W_{crit}$) |
| | | Case 3 | 382.43 | 383.98 | -1.55 | 1.55 | 8 | |
| | | Case 4 | 229.41 | 229.80 | -0.39 | 0.39 | 4.50 | |
| | | Case 5 | 292.34 | 293.28 | -0.94 | 0.94 | 7 | |
| | | Case 6 | 376.53 | 376.57 | -0.04 | 0.04 | 1 | |
| | | Case 7 | 207.56 | 207.93 | -0.37 | 0.37 | 3 | |
| | | Case 8 | 271.40 | 271.64 | -0.24 | 0.24 | 2 | |
| | | Case 9 | 378.02 | 378.77 | -0.75 | 0.75 | 6 | |
| | Mean Weight | Case 1 | 8640.82 | 8647.37 | -6.55 | 6.55 | 8 | 0 |
| | | Case 2 | 10709.61 | 10718.52 | -8.91 | 8.91 | 9 | ($<W_{crit}$) |
| | | Case 3 | 13323.81 | 13323.96 | -0.15 | 0.15 | 3 | |
| | | Case 4 | 8277.24 | 8277.76 | -0.52 | 0.52 | 4 | |
| | | Case 5 | 10087.33 | 10089.27 | -1.94 | 1.94 | 6 | |

|  |  | Case 6 | 13104.87 | 13105.60 | -0.73 | 0.73 | 5 |  |
|  |  | Case 7 | 8093.61 | 8093.70 | -0.09 | 0.09 | 1 |  |
|  |  | Case 8 | 9444.77 | 9449.88 | -5.11 | 5.11 | 7 |  |
|  |  | Case 9 | 13179.25 | 13179.35 | -0.10 | 0.10 | 2 |  |
|  | Mean $CO_2$ | Case 1 | 1518.73 | 1519.29 | -0.56 | 0.56 | 7 | 0 ($<W_{crit}$) |
|  |  | Case 2 | 1918.97 | 2001.08 | -82.11 | 82.11 | 9 |  |
|  |  | Case 3 | 2387.40 | 2387.49 | -0.09 | 0.09 | 2 |  |
|  |  | Case 4 | 1403.28 | 1403.30 | -0.02 | 0.02 | 1 |  |
|  |  | Case 5 | 1797.11 | 1797.48 | -0.37 | 0.37 | 4 |  |
|  |  | Case 6 | 2338.04 | 2338.12 | -1.08 | 1.08 | 8 |  |
|  |  | Case 7 | 1262.60 | 1262.98 | -0.38 | 0.38 | 5 |  |
|  |  | Case 8 | 1668.25 | 1668.66 | -0.41 | 0.41 | 6 |  |
|  |  | Case 9 | 2347.29 | 2347.64 | -0.35 | 0.35 | 3 |  |
| Example 1 | Mean Cost | Observations | FAGLSUD | PSO | Differences | Absolute differences | Rank | $W_{stat}$ |
|  |  | Case 1 | 62.45 | 62.86 | -0.41 | 0.41 | 5 | 0 ($<W_{crit}$) |
|  |  | Case 2 | 83.93 | 84.01 | -0.08 | 0.08 | 3 |  |
|  |  | Case 3 | 116.71 | 117.36 | -0.65 | 0.65 | 7 |  |
|  |  | Case 4 | 59.35 | 59.78 | -0.43 | 0.43 | 6 |  |
|  |  | Case 5 | 79.51 (79.5101) | 79.51 (79.5142) | -0.0041 | 0.0041 | 1 |  |
|  |  | Case 6 | 118.90 | 119.63 | -0.73 | 0.73 | 8 |  |
|  |  | Case 7 | 56.88 | 57.15 | -0.27 | 0.27 | 4 |  |
|  |  | Case 8 | 75.44 | 75.49 | -0.05 | 0.05 | 2 |  |
|  |  | Case 9 | 131.06 | 131.80 | -0.74 | 0.74 | 9 |  |
|  | Mean Weight | Case 1 | 2481.87 | 2487.09 | -5.22 | 5.22 | 5 | 0 ($<W_{crit}$) |
|  |  | Case 2 | 3055.33 | 3058.40 | -2.87 | 2.87 | 2 |  |
|  |  | Case 3 | 4208.16 | 4211.16 | -3.00 | 3.00 | 3 |  |
|  |  | Case 4 | 2472.46 | 2477.83 | -5.37 | 5.37 | 6 |  |
|  |  | Case 5 | 2950.91 | 2953.54 | -2.63 | 2.63 | 1 |  |
|  |  | Case 6 | 4308.41 | 4312.14 | -3.73 | 3.73 | 4 |  |
|  |  | Case 7 | 2465.38 | 3058.40 | -593.02 | 593.02 | 9 |  |
|  |  | Case 8 | 2858.19 | 3058.40 | -200.21 | 200.21 | 8 |  |
|  |  | Case 9 | 4781.25 | 4789.49 | -8.24 | 8.24 | 7 |  |
|  | Mean $CO_2$ | Case 1 | 379.97 | 380.88 | -0.91 | 0.91 | 3 | 0 ($<W_{crit}$) |
|  |  | Case 2 | 512.66 | 513.31 | -0.65 | 0.65 | 2 |  |
|  |  | Case 3 | 712.77 | 721.6 | -8.83 | 8.83 | 9 |  |
|  |  | Case 4 | 359.07 | 361.17 | -2.10 | 2.10 | 6 |  |
|  |  | Case 5 | 484.01 | 485.55 | -1.54 | 1.54 | 4 |  |
|  |  | Case 6 | 728.47 | 734.68 | -6.21 | 6.21 | 8 |  |
|  |  | Case 7 | 342.11 | 344.07 | -1.96 | 1.96 | 5 |  |
|  |  | Case 8 | 459.37 | 459.48 | -0.11 | 0.11 | 1 |  |
|  |  | Case 9 | 805.37 | 810.16 | -4.79 | 4.79 | 7 |  |
| Example 2 | Mean Cost | Case 1 | 247.39 | 247.85 | -0.46 | 0.46 | 4 | 0 ($<W_{crit}$) |
|  |  | Case 2 | 311.00 | 311.74 | -0.74 | 0.74 | 5 |  |
|  |  | Case 3 | 382.43 | 384.26 | -1.83 | 1.83 | 8 |  |
|  |  | Case 4 | 229.41 | 247.85 | -18.44 | 18.44 | 9 |  |
|  |  | Case 5 | 292.34 | 292.38 | -0.04 | 0.04 | 1 |  |
|  |  | Case 6 | 376.53 | 377.50 | -0.97 | 0.97 | 7 |  |
|  |  | Case 7 | 207.56 | 207.69 | -0.13 | 0.13 | 2 |  |
|  |  | Case 8 | 271.40 | 271.73 | -0.33 | 0.33 | 3 |  |
|  |  | Case 9 | 378.02 | 378.88 | -0.86 | 0.86 | 6 |  |
|  | Mean Weight | Case 1 | 8640.82 | 8644.37 | -3.55 | 3.55 | 7 | 7 ($>W_{crit}$) |
|  |  | Case 2 | 10709.61 | 10709.76 | -0.15 | 0.15 | 1 |  |

| | | Case 3 | 13323.81 | 13325.32 | -1.51 | 1.51 | 6 | |
|---|---|---|---|---|---|---|---|---|
| | | Case 4 | 8277.24 | 8281.45 | -4.21 | 4.21 | 8 | |
| | | Case 5 | 10087.33 | 10086.42 | +0.91 | 0.91 | 4 | |
| | | Case 6 | 13104.87 | 13105.54 | -0.67 | 0.67 | 2 | |
| | | Case 7 | 8093.61 | 8098.19 | -4.58 | 4.58 | 9 | |
| | | Case 8 | 9444.77 | 9444.05 | +0.72 | 0.72 | 3 | |
| | | Case 9 | 13179.25 | 13180.74 | -1.49 | 1.49 | 5 | |
| | **Mean $CO_2$** | Case 1 | 1518.73 | 1521.29 | -2.56 | 2.56 | 4 | 0 ($<W_{crit}$) |
| | | Case 2 | 1918.97 | 1923.36 | -4.39 | 4.39 | 6 | |
| | | Case 3 | 2387.40 | 2392.94 | -5.54 | 5.54 | 8 | |
| | | Case 4 | 1403.28 | 1404.99 | -1.71 | 1.71 | 1 | |
| | | Case 5 | 1797.11 | 1799.73 | -2.62 | 2.62 | 5 | |
| | | Case 6 | 2338.04 | 2342.87 | -4.83 | 4.83 | 7 | |
| | | Case 7 | 1262.60 | 1265.12 | -2.52 | 2.52 | 3 | |
| | | Case 8 | 1668.25 | 1670.51 | -2.26 | 2.26 | 2 | |
| | | Case 9 | 2347.29 | 2353.55 | -6.26 | 6.26 | 9 | |
| **Example 1** | **Mean Cost** | **Observations** | **FAGLSUD** | **LFBBO** | **Differences** | **Absolute differences** | **Rank** | **$W_{stat}$** |
| | | Case 1 | 62.45 | 63.16 | -0.71 | 0.71 | 3 | 0 ($<W_{crit}$) |
| | | Case 2 | 83.93 | 89.08 | -5.15 | 5.15 | 9 | |
| | | Case 3 | 116.71 | 121.31 | -4.60 | 4.60 | 8 | |
| | | Case 4 | 59.35 | 59.37 | -0.02 | 0.02 | 2 | |
| | | Case 5 | 79.51 | 82.33 | -2.82 | 2.82 | 7 | |
| | | Case 6 | 118.90 | 121.33 | -2.43 | 2.43 | 6 | |
| | | Case 7 | 56.88 | 56.89 | -0.01 | 0.01 | 1 | |
| | | Case 8 | 75.44 | 77.21 | -1.77 | 1.77 | 5 | |
| | | Case 9 | 131.06 | 131.83 | -0.77 | 0.77 | 4 | |
| | **Mean Weight** | Case 1 | 2481.87 | 2481.96 | -0.09 | 0.09 | 2 | 0 ($<W_{crit}$) |
| | | Case 2 | 3055.33 | 3124.30 | -68.97 | 68.97 | 6 | |
| | | Case 3 | 4208.16 | 4367.50 | -159.34 | 159.34 | 8 | |
| | | Case 4 | 2472.46 | 2472.63 | -0.17 | 0.17 | 3 | |
| | | Case 5 | 2950.91 | 2952.64 | -1.73 | 1.73 | 5 | |
| | | Case 6 | 4308.41 | 4520.78 | -212.37 | 212.37 | 9 | |
| | | Case 7 | 2465.38 | 2465.45 | -0.07 | 0.07 | 1 | |
| | | Case 8 | 2858.19 | 2858.56 | -0.37 | 0.37 | 4 | |
| | | Case 9 | 4781.25 | 4876.12 | -94.87 | 94.87 | 7 | |
| | **Mean $CO_2$** | Case 1 | 379.97 | 386.36 | -6.39 | 6.39 | 4 | 0 ($<W_{crit}$) |
| | | Case 2 | 512.66 | 556.09 | -43.43 | 43.43 | 8 | |
| | | Case 3 | 712.77 | 800.85 | -88.08 | 88.08 | 9 | |
| | | Case 4 | 359.07 | 359.96 | -0.89 | 0.89 | 2 | |
| | | Case 5 | 484.01 | 510.76 | -26.75 | 26.75 | 7 | |
| | | Case 6 | 728.47 | 741.45 | -12.98 | 12.98 | 5 | |
| | | Case 7 | 342.11 | 342.18 | -0.07 | 0.07 | 1 | |
| | | Case 8 | 459.37 | 473.40 | -14.03 | 14.03 | 6 | |
| | | Case 9 | 805.37 | 808.68 | -3.31 | 3.31 | 3 | |
| **Example 2** | **Mean Cost** | Case 1 | 247.39 | 248.33 | -0.94 | 0.94 | 3 | 0 ($<W_{crit}$) |
| | | Case 2 | 311.00 | 315.27 | -4.27 | 4.27 | 9 | |
| | | Case 3 | 382.43 | 385.75 | -3.32 | 3.32 | 8 | |
| | | Case 4 | 229.41 | 229.42 | -0.01 | 0.01 | 1 | |
| | | Case 5 | 292.34 | 294.03 | -1.69 | 1.69 | 4 | |
| | | Case 6 | 376.53 | 378.33 | -1.80 | 1.80 | 5 | |
| | | Case 7 | 207.56 | 207.84 | -0.28 | 0.28 | 2 | |
| | | Case 8 | 271.40 | 274.32 | -2.92 | 2.92 | 7 | |
| | | Case 9 | 378.02 | 380.09 | -2.07 | 2.07 | 6 | |
|  | **Mean Weight** | Case 1 | 8640.82 | 9091.10 | -450.28 | 450.28 | 7 | 0 ($<W_{crit}$) |
|  |  | Case 2 | 10709.61 | 10744.01 | -34.40 | 34.40 | 2 |  |
|  |  | Case 3 | 13323.81 | 13439.86 | -116.05 | 116.05 | 5 |  |
|  |  | Case 4 | 8277.24 | 9024.82 | -747.58 | 747.58 | 8 |  |
|  |  | Case 5 | 10087.33 | 10186.39 | -132.06 | 132.06 | 6 |  |
|  |  | Case 6 | 13104.87 | 13196.88 | -92.01 | 92.01 | 4 |  |
|  |  | Case 7 | 8093.61 | 8093.70 | -0.09 | 0.09 | 1 |  |
|  |  | Case 8 | 9444.77 | 10486.56 | -1041.79 | 1041.79 | 9 |  |
|  |  | Case 9 | 13179.25 | 13260.20 | -80.95 | 80.95 | 3 |  |
|  | **Mean $CO_2$** | Case 1 | 1518.73 | 1550.47 | -31.74 | 31.74 | 8 | 0 ($<W_{crit}$) |
|  |  | Case 2 | 1918.97 | 1946.65 | -27.68 | 27.68 | 6 |  |
|  |  | Case 3 | 2387.40 | 2399.99 | -12.59 | 12.59 | 5 |  |
|  |  | Case 4 | 1403.28 | 1403.73 | -0.45 | 0.45 | 1 |  |
|  |  | Case 5 | 1797.11 | 1856.41 | -59.30 | 59.30 | 9 |  |
|  |  | Case 6 | 2338.04 | 2366.00 | -27.96 | 27.96 | 7 |  |
|  |  | Case 7 | 1262.60 | 1273.20 | -10.60 | 10.60 | 3 |  |
|  |  | Case 8 | 1668.25 | 1669.86 | -1.61 | 1.61 | 2 |  |
|  |  | Case 9 | 2347.29 | 2358.57 | -11.28 | 11.28 | 4 |  |

Here, we provide a detailed pairwise comparison between FAGLSUD and the other RCC design optimizers used in this study (MSA, PSO, and LFBBO) using all six sets of case observations.

For a pairwise comparison between FAGLSUD and MSA, we calculated the following quantities: Rank-sum for negative differences $T_- = 45$, Rank-sum for positive differences $T_+ = 0$, Wilcoxon rank signed rank test statistic $W_{stat} = \min(T_-, T_+) = 0$, and Critical value $W_{crit} = 5$ at $\alpha = 0.05$. The obtained relation $W_{stat} < W_{crit}$ confirms that the median differences between FAGLSUD and MSA in all observations and over all objectives are zero. Hence, the null hypothesis $H_0$ is rejected in favor of the alternative hypothesis $H_1$ that the median differences are not zero. Therefore, we have found a significant difference between FAGLSUD and MSA.

Based on the same quantities as above, we can see that FAGLSUD is statistically superior to LFBBO and other less competitive methods for all six sets of case observations. The obtained relation $W_{stat} < W_{crit}$ in all case observations and across all objective values confirms that FAGLSUD is statistically significantly superior to the other methods.

The only exception is the pairwise comparison between FAGLSUD and PSO, where FAGLSUD is statistically superior to PSO in only 5 of the 6 sets of case observations. For 1 set of case observations, we obtained the relation $W_{stat} > W_{crit}$, which indicates that the median differences between FAGLSUD and PSO (mean weight objective values in example 2) are nonzero; and thus, both perform statistically similar (and superior to other methods). The measured quantities are: Rank-sum for negative differences $T_- = 38$, Rank-sum for positive differences $T_+ = 7$, Wilcoxon rank signed rank test statistic $W_{stat} = \min(T_-, T_+) = 7$, and Critical value $W_{crit} = 5$ at $\alpha = 0.05$.